\documentclass[10pt]{article}
\usepackage[preprint]{tmlr}
\usepackage{amsmath,amssymb,graphicx,booktabs,tabularx}
\usepackage{microtype}
\usepackage{xurl}
\usepackage{float}
\usepackage{placeins}
\usepackage{hyperref}
\usepackage{url}
\usepackage[utf8]{inputenc}
\hypersetup{pdftitle={From Data to Wisdom: Rethinking AI Through an Epistemic Lens},pdfauthor={Afonso Eduardo},pdfkeywords={Artificial Intelligence, Machine Learning, Epistemology, Knowledge Management, Epistemic Architecture, DIKUW, Bayesian Reasoning, Model Evaluation, Agentic AI}}
\graphicspath{{figures/}}
\title{From Data to Wisdom: Rethinking AI Through an Epistemic Lens}
\author{\name Afonso Eduardo \email aflmeduardo@gmail.com}
\begin{document}
\maketitle

\begin{abstract}
This paper proposes the Data-to-Wisdom Epistemic Architecture for AI (EAAI), a conceptual framework treating Data, Information, Knowledge, Understanding and Wisdom as five distinct but recursively interacting epistemic functions: observation, contextual interpretation, modeling and prediction, critical evaluation and informed judgment. Two observations motivate it: sequential readings of the Knowledge Pyramid obscure how these functions interact, and a targeted review of twelve Bayesian reference books finds that prior distributions are defined mathematically while ``prior information'' and ``prior knowledge'' are not consistently distinguished as epistemic categories. A Bayesian interpretation relates the likelihood, prior and posterior distributions, posterior prediction, model checking and decision analysis to the five functions, without limiting EAAI to Bayesian systems. Extending ``prior'' in an epistemic sense, the paper distinguishes five kinds of prior context that can shape the model, the likelihood, the prior, evaluation and action. Three worked applications---quantitative financial risk, actuarial modeling in health insurance and agentic AI for medical reimbursement claims---show how EAAI structures the analysis of model assumptions, evidence interpretation, validation, dependencies and decision consequences, separating the prediction of a risk, the appraisal of its reliability and the justification of action. It draws implications for AI and machine learning, including evaluation by function, treating intermediate model outputs as evidence open to challenge and separating generation from appraisal, and compares EAAI with work on DIKW, Bayesian workflow, AI evaluation and risk governance. Its contributions are an epistemic vocabulary with recursive relationships, a more accessible reading of Bayesian methodology, a typology of prior context and clearer methodological terminology.
\end{abstract}
\vspace{8pt}
\section{Introduction}
This article introduces the Data-to-Wisdom Epistemic Architecture for AI (EAAI), a coherent conceptual framework for examining the epistemic functions involved in artificial intelligence and machine learning, including their interactions, dependencies and recursive relationships. Building on the five categories of Data, Information, Knowledge, Understanding and Wisdom, EAAI proposes that these functions are distinct but interconnected, rather than stages of a mandatory progression. The architecture provides a perspective for examining how AI systems interpret evidence, develop and use models, generate predictions, evaluate their reliability and inform decisions. These functions are performed not only by AI systems but also by the people and institutions that build, evaluate and use them. Bayesian methodology offers a particularly illuminating interpretive lens, but the architecture applies more broadly to AI and machine learning.

The article makes four contributions. First, it defines Data, Information, Knowledge, Understanding and Wisdom as epistemic functions with explicit recursive relationships, providing a shared epistemic vocabulary for AI research and practice (Section~\ref{sec:eaai}). Second, it maps Bayesian methodology onto these functions, making the purpose of its tools accessible to readers who do not use Bayesian terminology (Section~\ref{sec:bayes}). Third, it distinguishes five kinds of prior context and shows that they can shape the choice of model and likelihood, the prior, evaluation and action, not only the prior distribution (Section~\ref{sec:prior-context}). Fourth, it clarifies methodology and terminology that the examined literature leaves ambiguous, including prior information, prior knowledge and evidence, and the separation of prediction, evaluation and decision (Sections~\ref{sec:bayes-books} and~\ref{sec:examples}).

\begin{figure}[p]
\centering\includegraphics[width=\linewidth,height=.92\textheight,keepaspectratio]{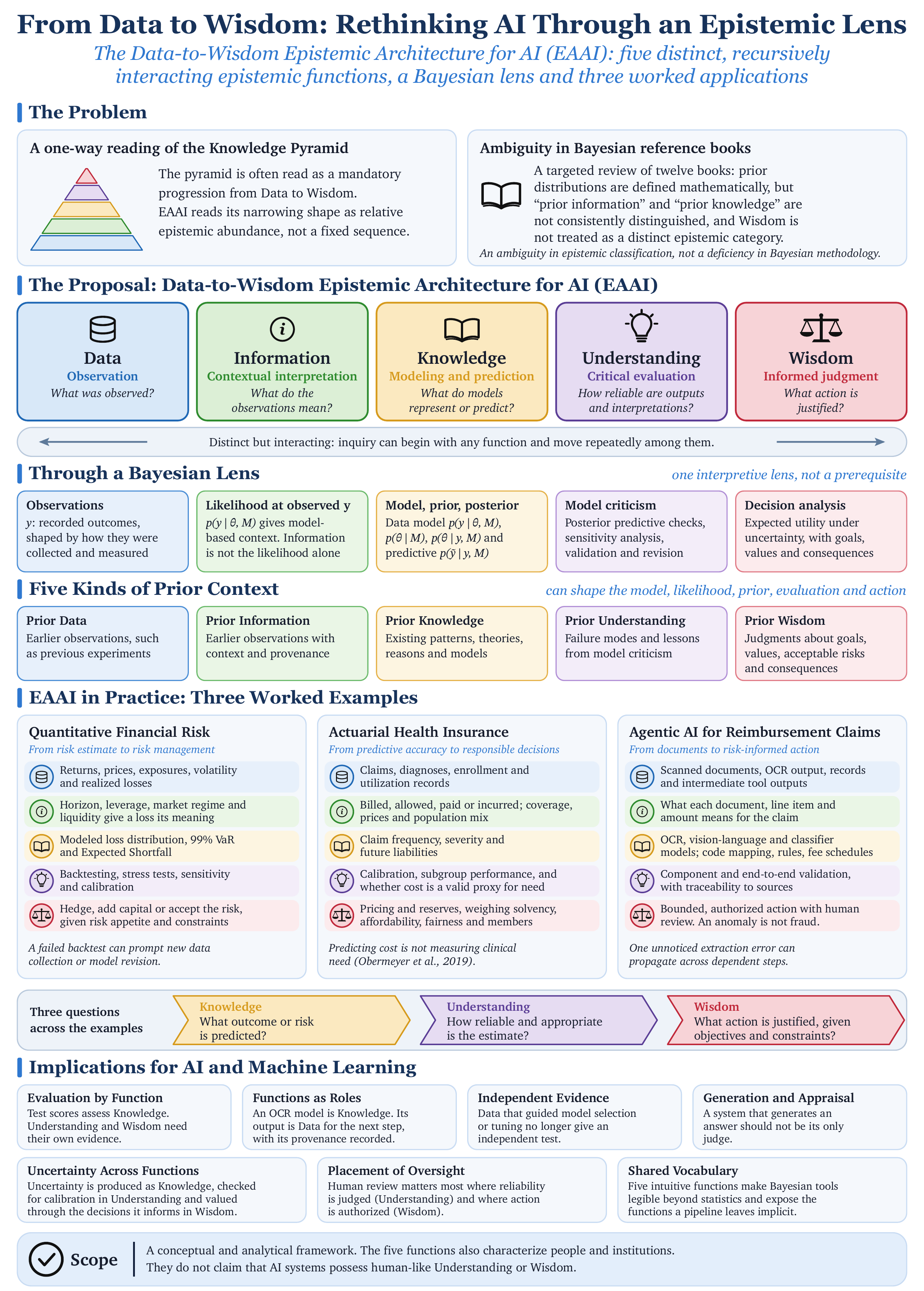}
\caption{Overview of the article: the problem, the Data-to-Wisdom Epistemic Architecture for AI (EAAI), its Bayesian interpretation, five kinds of prior context, three worked examples and implications for AI and machine learning.}
\label{fig:overview}
\end{figure}

The Knowledge Pyramid, long discussed in information science, is often depicted as a progression from Data through Information, Knowledge and Understanding to Wisdom. This article interprets its narrowing shape in terms of relative epistemic abundance, while EAAI addresses a different question: how the five functions interact throughout inquiry. The discussion moves from the framework’s intellectual foundations and a targeted review of Bayesian terminology to the general architecture, its Bayesian interpretation and practical applications in quantitative risk assessment, actuarial modeling and agentic AI. Figure~\ref{fig:overview} summarizes the article in a single infographic. Near the end, Section~\ref{sec:related} compares EAAI with related work: interpretations and critiques of the DIKW hierarchy, philosophy of understanding, Bayesian workflow, cognitive architectures, AI evaluation and risk governance.

\section{Origins and the Problem with a One-Way Knowledge Pyramid}
An earlier literary precursor appears in T. S. Eliot’s Choruses from “The Rock” (1934). Eliot asks, “Where is the wisdom we have lost in knowledge? Where is the knowledge we have lost in information?” His lines distinguish information, knowledge and wisdom more than five decades before Ackoff’s (1989) five-category formulation. Eliot’s poetic question was not a formal taxonomy, and the article makes no claim that Ackoff derived his classification from Eliot.

\begin{figure}[tbp]
\centering\includegraphics[width=\linewidth,height=.63\textheight,keepaspectratio]{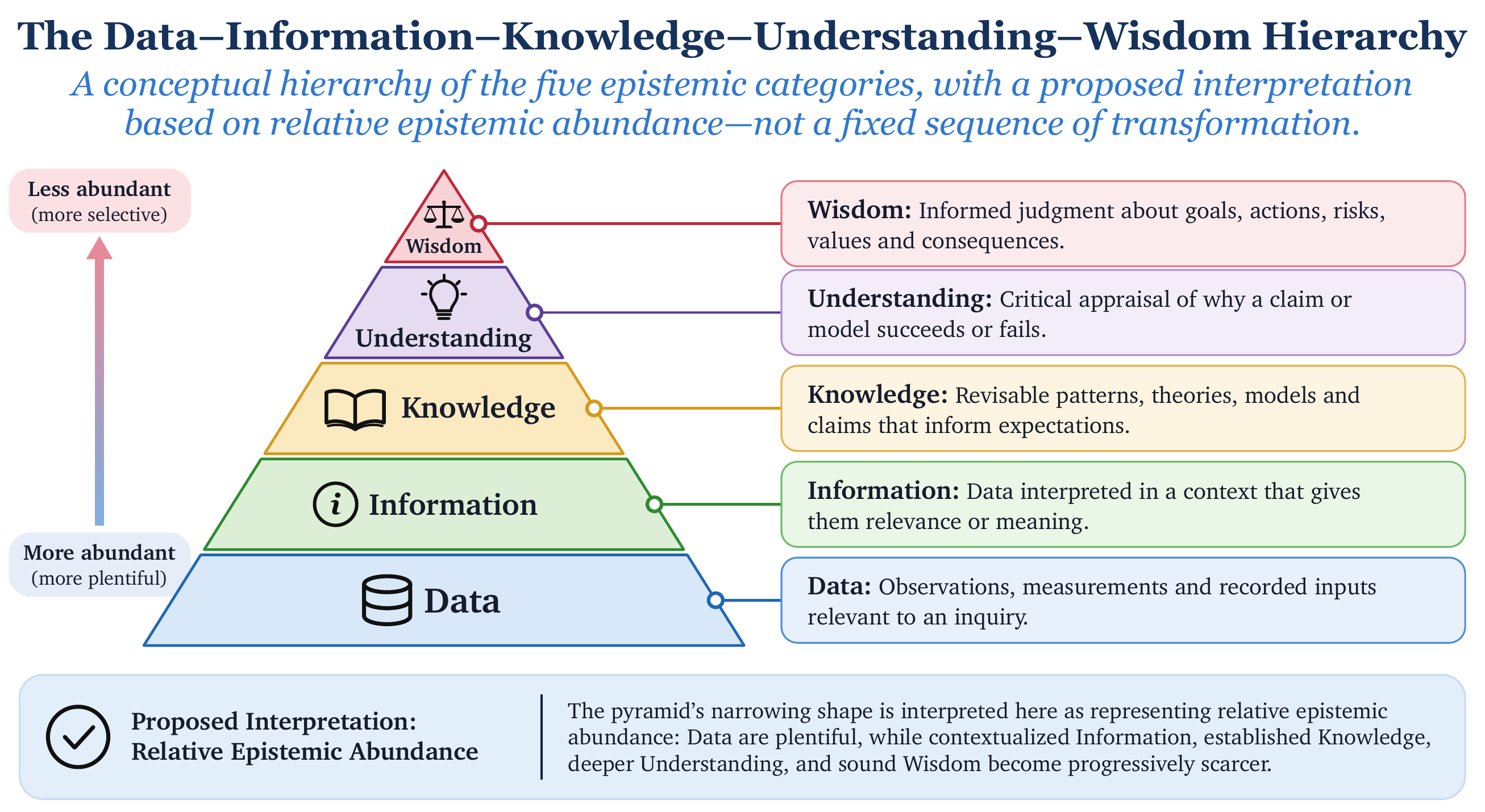}
\caption{The DIKUW hierarchy interpreted through relative epistemic abundance, drawing on the five categories of Ackoff (1989).}
\label{fig:pyramid}
\end{figure}

The data–information–knowledge–wisdom (DIKW) hierarchy is a model used widely in information science and knowledge management, usually depicted as a pyramid with data at its base and wisdom at its apex (Frické, 2019). Russell Ackoff’s “From Data to Wisdom” (1989) is an influential formulation that included a fifth level, Understanding, between Knowledge and Wisdom. Later versions of the hierarchy typically omit it (Frické, 2019). Ackoff described the hierarchy in prose, and its depiction as a pyramid came later. This article restores Understanding, because the four-level DIKW hierarchy leaves no place for the critical evaluation of models and claims, a function central to AI and machine learning. “DIKUW” is used here as shorthand for Ackoff’s five categories. Figure~\ref{fig:pyramid} shows the hierarchy with the interpretation proposed in this article.

This article interprets the pyramid’s narrowing shape as a representation of relative epistemic abundance. Data may be plentiful, while contextualized Information, established Knowledge, deeper Understanding and sound Wisdom become progressively scarcer. This concerns the relative availability of epistemic resources rather than a sequence through which one category must be transformed into another.

Rowley (2007) documented inconsistencies in how the DIKW hierarchy defines and relates data, information, knowledge and wisdom. Frické (2009) challenged the assumption that knowledge arises through a straightforward progression from uninterpreted data. Both critiques question the treatment of epistemic categories as fixed stages of an upward process. Observations, instead, are selected and interpreted in relation to questions, theories, assumptions and context.

Relative abundance alone does not explain how epistemic functions interact. A scientist may begin with a surprising observation, a policymaker with a decision or a statistician with an existing model. Each can move recursively among evidence, interpretation, models, critical appraisal and judgment. These relationships motivate the general architecture developed below.

Ackoff’s (1989) five-category distinction supplies the historical vocabulary. EAAI develops these categories into an interconnected epistemic architecture with a decision-theoretic dimension.

\section{Epistemic Ambiguities in Bayesian Reference Books: Information, Knowledge and Wisdom}\label{sec:bayes-books}
The preceding discussion distinguishes five epistemic categories and proposes an interpretation of their relative abundance. A related question arises when these categories are examined in Bayesian modeling: what exactly is meant by prior Information or prior Knowledge and how are these distinguished from a prior probability distribution?

A targeted textual review examined twelve books spanning Bayesian statistics, statistical decision theory, probabilistic machine learning, Bayesian optimization and decision making under uncertainty. The works examined were Bayesian Data Analysis (Gelman et al., 2013); Bayesian Workflow (Gelman et al., 2026); Statistical Decision Theory and Bayesian Analysis (Berger, 1985); Information Theory, Inference, and Learning Algorithms (MacKay, 2003); Gaussian Processes for Machine Learning (Rasmussen and Williams, 2006); Pattern Recognition and Machine Learning (Bishop, 2006); Probabilistic Machine Learning: An Introduction and Probabilistic Machine Learning: Advanced Topics (Murphy, 2022, 2023); Bayesian Reasoning and Machine Learning (Barber, 2012); Bayesian Inference: Theory, Methods, Computations (Zwanzig and Ahmad, 2024); Bayesian Optimization (Garnett, 2023); and Algorithms for Decision Making (Kochenderfer et al., 2022). For this article, the author used frontier AI tools to support searches for terminology and examination of passages in accessible editions, with interpretations reviewed in context. This is a targeted textual analysis, not an exhaustive survey of Bayesian scholarship. Appendix~\ref{app:review} gives the editions, the passages with their pages and the counts of each term.

\begin{figure}[tbp]
\centering\includegraphics[width=\linewidth,height=.63\textheight,keepaspectratio]{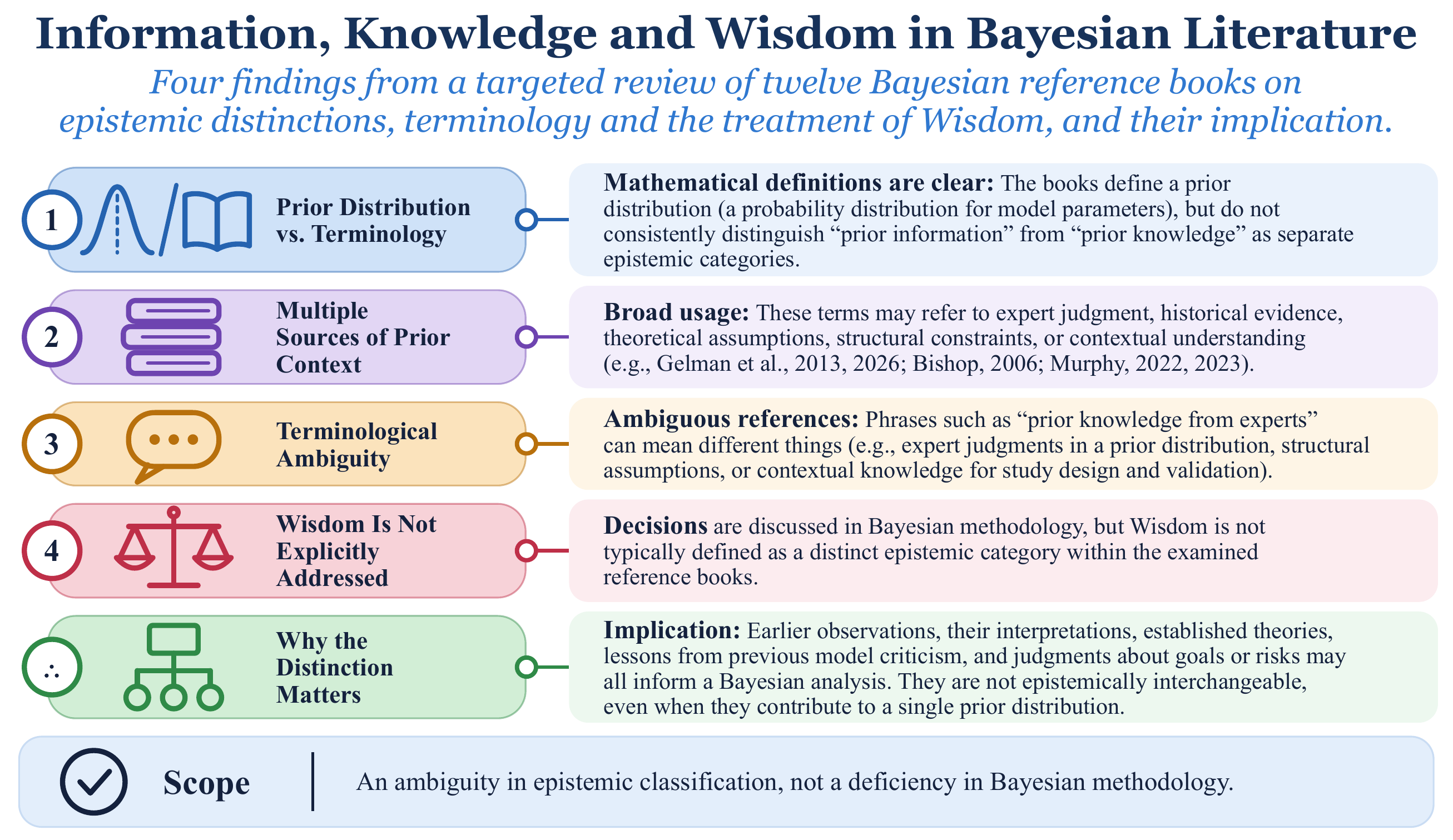}
\caption{Information, Knowledge and Wisdom in Bayesian literature: four findings from a targeted review of twelve reference books on epistemic distinctions, terminology and the treatment of Wisdom, and their implication.}
\label{fig:bayes-lit}
\end{figure}

The central observation from this targeted review is that the examined books define prior distributions mathematically but do not consistently establish “prior information” and “prior knowledge” as distinct epistemic categories. These expressions may encompass expert judgment, historical evidence, theoretical assumptions, structural constraints or contextual understanding (Gelman et al., 2013, 2026; Bishop, 2006; Murphy, 2022, 2023). The observation is limited to the passages and editions examined. It does not suggest that Bayesian authors neglect prior knowledge or that Bayesian methodology lacks formal mechanisms for incorporating it. Figure~\ref{fig:bayes-lit} summarizes four findings and their implication.

The distinction matters because different forms of prior context play different roles in inquiry. Earlier observations, interpretations of those observations, established theories, lessons from previous model criticism and judgments about goals or risks may all inform a Bayesian analysis. Yet they are not epistemically interchangeable, even when several contribute to the specification of a single prior distribution.

Consider a hypothetical statement: “The model incorporates prior knowledge from experts.” This could describe expert judgments encoded in a probability distribution, structural assumptions built into the model or contextual knowledge informing study design and validation. Each possibility has different implications for transparency, evaluation and subsequent revision. The terminology alone does not reveal which role is intended.

A related distinction concerns the word “evidence.” In Bayesian statistics, model evidence can refer specifically to the marginal likelihood, $p(y\mid M)$. In broader epistemic language, evidence refers to observations or reasons supporting a claim. These meanings should not be conflated and EAAI does not identify Information with the technical quantity called Bayesian evidence (MacKay, 2003; Murphy, 2022).

The gap identified in this targeted review concerns epistemic articulation rather than a missing statistical operation. Bayesian methodology already offers rigorous tools for inference, prediction, model criticism and decision analysis. The observation nevertheless raises a broader question: how can the roles of observations, contextual interpretation, models, critical evaluation and informed judgment be distinguished while recognizing their interactions throughout AI research and practice? The following section introduces EAAI as a general epistemic architecture addressing that question, with Bayesian methodology subsequently serving as one interpretive lens.

\section{Data-to-Wisdom Epistemic Architecture for AI (EAAI)}\label{sec:eaai}
\subsection{EAAI as a General Epistemic Architecture}
The Data-to-Wisdom Epistemic Architecture for AI (EAAI) organizes Data, Information, Knowledge, Understanding and Wisdom as distinct, interacting functions. Whereas the earlier pyramid interpretation concerns relative epistemic abundance, EAAI describes how the functions influence one another during evidence collection, interpretation, modeling, prediction, critical evaluation and decision-making.

This perspective applies to Bayesian and non-Bayesian AI, including systems whose development, evaluation and use involve human judgment and oversight.

\begin{figure}[tbp]
\centering
\includegraphics[width=\linewidth,height=.63\textheight,keepaspectratio]{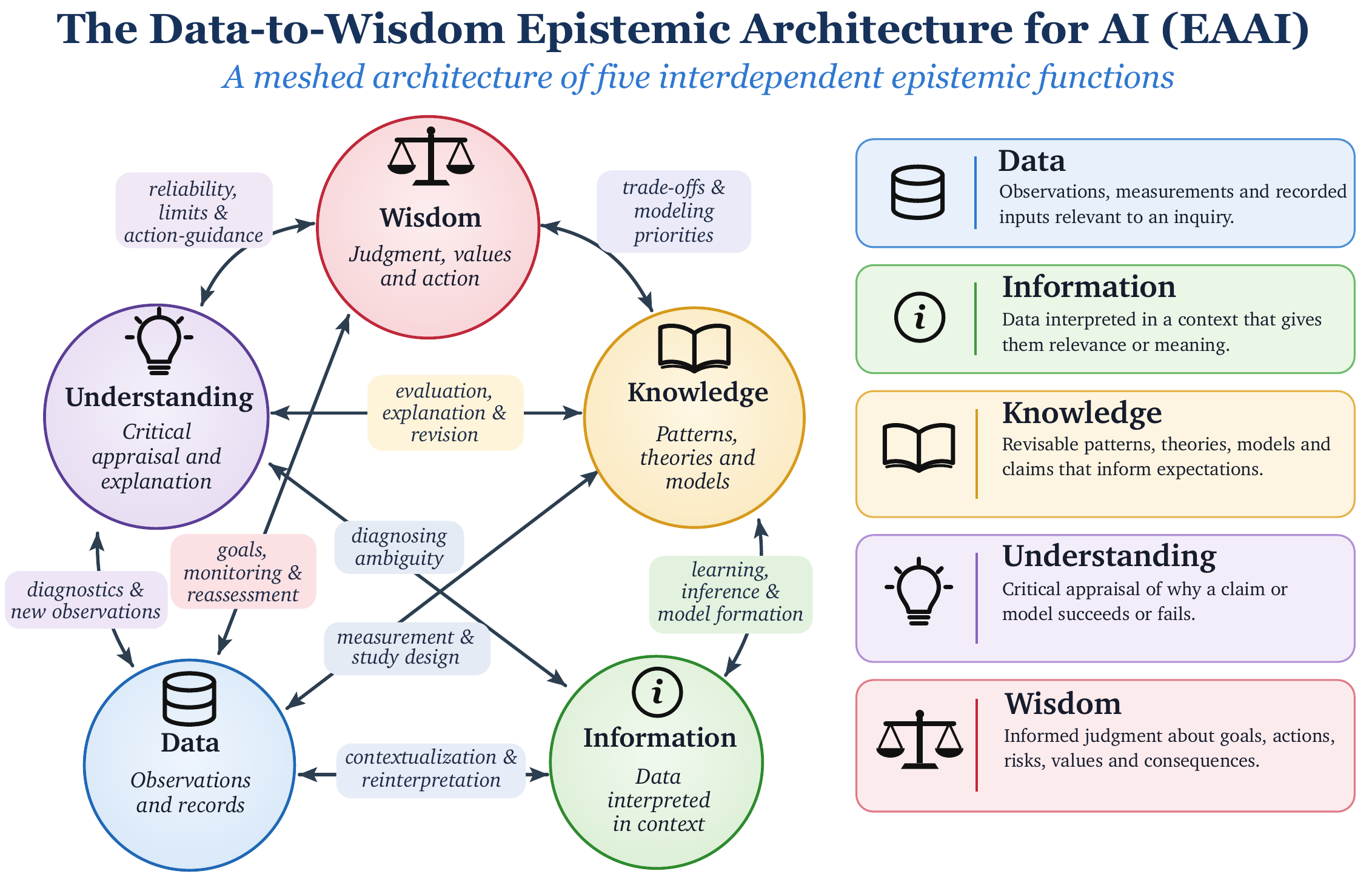}
\caption{The Data-to-Wisdom Epistemic Architecture for AI (EAAI). The labeled reciprocal and cross-functional connections illustrate how context, inference, model evaluation, decision requirements and reassessment link the five functions. The relationships are illustrative rather than exhaustive.}
\label{fig:eaai-network}
\end{figure}

The five categories of Data, Information, Knowledge, Understanding and Wisdom build on Ackoff's (1989) formulation. This paper extends their interpretation by defining them as interacting epistemic functions within EAAI. The proposed architecture emphasizes their distinct roles, dependencies and recursive relationships across AI research and practice, including modeling, prediction, evaluation and decision-making. Table~\ref{tab:eaai-functions} presents their definitions in this context, while Figure~\ref{fig:eaai-network} illustrates selected interactions among the functions.

\begin{table}[tbp]
\centering
\caption{The five epistemic functions of EAAI. Definitions adapted and extended from the five-category framework discussed by Ackoff (1989) for the analysis of AI and machine learning.}
\label{tab:eaai-functions}
\small
\renewcommand{\arraystretch}{1.18}
\begin{tabularx}{\linewidth}{@{}>{\raggedright\arraybackslash}p{0.21\linewidth}>{\raggedright\arraybackslash}X@{}}
\toprule
\textbf{Function} & \textbf{Description} \\
\midrule
\textbf{Data} & Observations, measurements and recorded inputs relevant to an inquiry. The choice of observations and of how to measure them depends on the questions and practices of the study. \\
\addlinespace
\textbf{Information} & Data interpreted within a context that gives them relevance or meaning, including the task, measurement definitions, provenance and study design. \\
\addlinespace
\textbf{Knowledge} & Patterns, theories, models and claims that inform expectations and can be supported, challenged or updated in light of evidence. A model or prediction is not automatically warranted Knowledge. \\
\addlinespace
\textbf{Understanding} & Critical and explanatory appraisal of relationships, mechanisms, assumptions and limitations, including examination of why a claim or model succeeds or fails. It involves more than producing a prediction. \\
\addlinespace
\textbf{Wisdom} & Informed judgment concerning goals, appropriate actions, trade-offs, risks, constraints, values and consequences. It is not synonymous with merely selecting an action or optimizing a numerical target. \\
\bottomrule
\end{tabularx}
\end{table}

These functions are recursive rather than sequential. Existing Knowledge may shape which Data are collected and how they become Information. Critical investigation can challenge earlier Knowledge, deepen Understanding or prompt new observations. Judgments about acceptable risks or consequences can reshape the original question or the decision to gather more evidence. Inquiry may therefore begin with any of the five functions and move repeatedly among them.

The functions answer different questions about the same application: What was observed? What do those observations mean? What do models represent or predict? How reliable are their outputs and interpretations? What action is justified? These questions can be revisited as new evidence or decision requirements emerge.

Figure~\ref{fig:eaai-network} presents these relationships as a reciprocal network, illustrating contextual interpretation, model development, critical evaluation, decision-guided evidence collection and revision.

\subsection{Prediction, Evaluation and Decision-Making Within EAAI}
Prediction is explicitly included within EAAI. A trained predictive model can encode learned relationships and generate estimates about uncertain outcomes, contributing to Knowledge. However, producing a prediction differs from examining its reliability, which belongs to Understanding, and from judging whether and how to act on it, which belongs to Wisdom. Data and Information remain essential throughout.

A model may perform well on average while becoming unreliable under distribution shift, exhibiting poor calibration for particular groups or failing under extreme conditions. Moreover, accurate prediction of an observable target does not automatically establish a causal explanation or show that the target represents the concept relevant to a decision. Even well-supported predictions do not uniquely determine an appropriate action: objectives, acceptable risks, constraints, alternatives and consequences matter.

\section{EAAI Through a Bayesian Lens}\label{sec:bayes}
The preceding general architecture applies across AI and machine learning. Bayesian methodology provides a particularly informative interpretation because it makes modeling assumptions, uncertainty, belief updating, prediction, model criticism and decisions explicit. The correspondences in Table~\ref{tab:bayes} illustrate how Bayesian tools can support the five functions. They neither exhaust those functions nor restrict EAAI to Bayesian approaches.

\begin{table}[!hbp]
\centering
\caption{EAAI through a Bayesian lens: Bayesian tools that can support each epistemic function. The correspondences are illustrative. They neither exhaust the functions nor restrict EAAI to Bayesian approaches.}
\label{tab:bayes}
\small
\renewcommand{\arraystretch}{1.18}
\begin{tabularx}{\linewidth}{@{}>{\raggedright\arraybackslash}p{0.15\linewidth}>{\raggedright\arraybackslash}p{0.22\linewidth}>{\raggedright\arraybackslash}X@{}}
\toprule
\textbf{Function} & \textbf{Bayesian correspondence} & \textbf{Description} \\
\midrule
\textbf{Data} & Recorded outcomes $y$ & Recorded outcomes $y$ supply the observations to be analyzed, with their meaning also shaped by how they were collected and measured. \\
\addlinespace
\textbf{Information} & Likelihood $p(y\mid\theta,M)$ at the observed $y$ & Evaluated at the observed outcomes $y$ and read as a function of $\theta$, the likelihood supplies model-based context for interpreting observations, contextualizing Data as Information within EAAI. Measurement definitions, study design and background assumptions also provide context. Information is not reducible to the likelihood alone. \\
\addlinespace
\textbf{Knowledge} & Data model $p(y\mid\theta,M)$, prior $p(\theta\mid M)$ and posterior $p(\theta\mid y,M)$ & The data model $p(y\mid\theta,M)$, specified as a distribution of $y$ for each $\theta$, and the prior distribution $p(\theta\mid M)$ formally represent modeling assumptions. The posterior distribution $p(\theta\mid y,M)$ represents their updating in light of observations. None of these distributions is identical to the full body of prior or acquired Knowledge. \\
\addlinespace
\textbf{Understanding} & Model checking and criticism & Model checking, posterior predictive assessment, sensitivity analysis, validation, causal investigation where appropriate and model revision can support critical appraisal of explanations and assumptions. These practices contribute to Understanding without constituting all of it. \\
\addlinespace
\textbf{Wisdom} & Decision analysis & Decision analysis can compare possible actions using expected utilities or losses under uncertainty, alongside risks and constraints. The epistemic function of Wisdom also requires examining the goals, values and consequences underlying those choices. Wisdom is not identical to maximizing expected utility. \\
\bottomrule
\end{tabularx}
\end{table}

The same expression $p(y\mid\theta,M)$ appears under two functions. As a specified data model, a distribution of $y$ for each $\theta$, it is a modeling assumption and part of Knowledge: Gelman et al.\ (2026) state that ``the specification of the data model represents prior information or assumptions''. Evaluated at the observed $y$ and read as a function of $\theta$, it is what Gelman et al.\ (2013) call the likelihood function, and it contextualizes Data as Information. The two readings are an instance of the recursive relationships of EAAI, not a conflict between functions.

\begin{figure}[tbp]
\centering\includegraphics[width=\linewidth,height=.63\textheight,keepaspectratio]{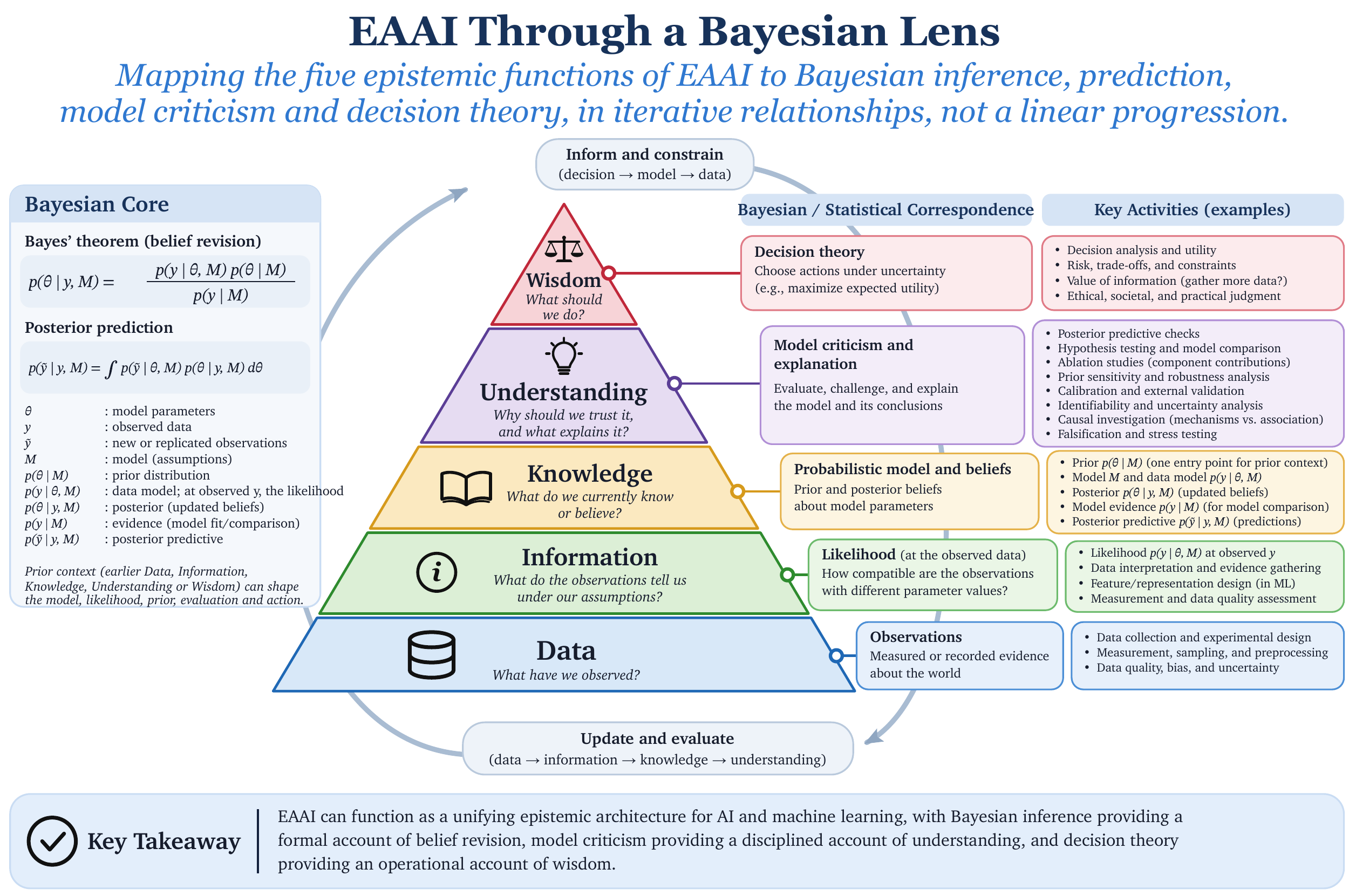}
\caption{EAAI viewed through a Bayesian lens: five epistemic functions (Data, Information, Knowledge, Understanding and Wisdom). The pyramid is a visual reference to the historical categories, not a mandatory progression or a complete depiction of the framework’s recursive relationships.}
\label{fig:bayes-mapping}
\end{figure}

Bayes' theorem makes belief revision explicit:
\begin{equation}\label{eq:bayes}
p(\theta\mid y,M)=\frac{p(y\mid\theta,M)\,p(\theta\mid M)}{p(y\mid M)}.
\end{equation}
Yet a posterior distribution, even when well computed, does not by itself establish an adequate explanation, a warranted causal claim or a justified decision. Model criticism and decision analysis address additional questions about reliability and action, while EAAI clarifies the distinct epistemic functions involved. Posterior prediction is represented by the posterior predictive distribution
\begin{equation}\label{eq:predictive}
p(\tilde y\mid y,M)=\int p(\tilde y\mid\theta,M)\,p(\theta\mid y,M)\,d\theta.
\end{equation}
Its predictions can be evaluated for calibration, generalization and decision relevance. Figure~\ref{fig:bayes-mapping} summarizes these correspondences.

The distinction between the general architecture and its Bayesian interpretation also matters for prior context. Earlier observations, contextualized Information, existing Knowledge, lessons contributing to Understanding and judgments involving Wisdom may all be available before an analysis begins. Several can inform the same statistical model without being interchangeable or reducible to a single prior distribution. The following workflow makes the iterative role of those resources concrete.

\section{Scientific Research and Bayesian Workflow}
\begin{figure}[tbp]
\centering\includegraphics[width=\linewidth,height=.63\textheight,keepaspectratio]{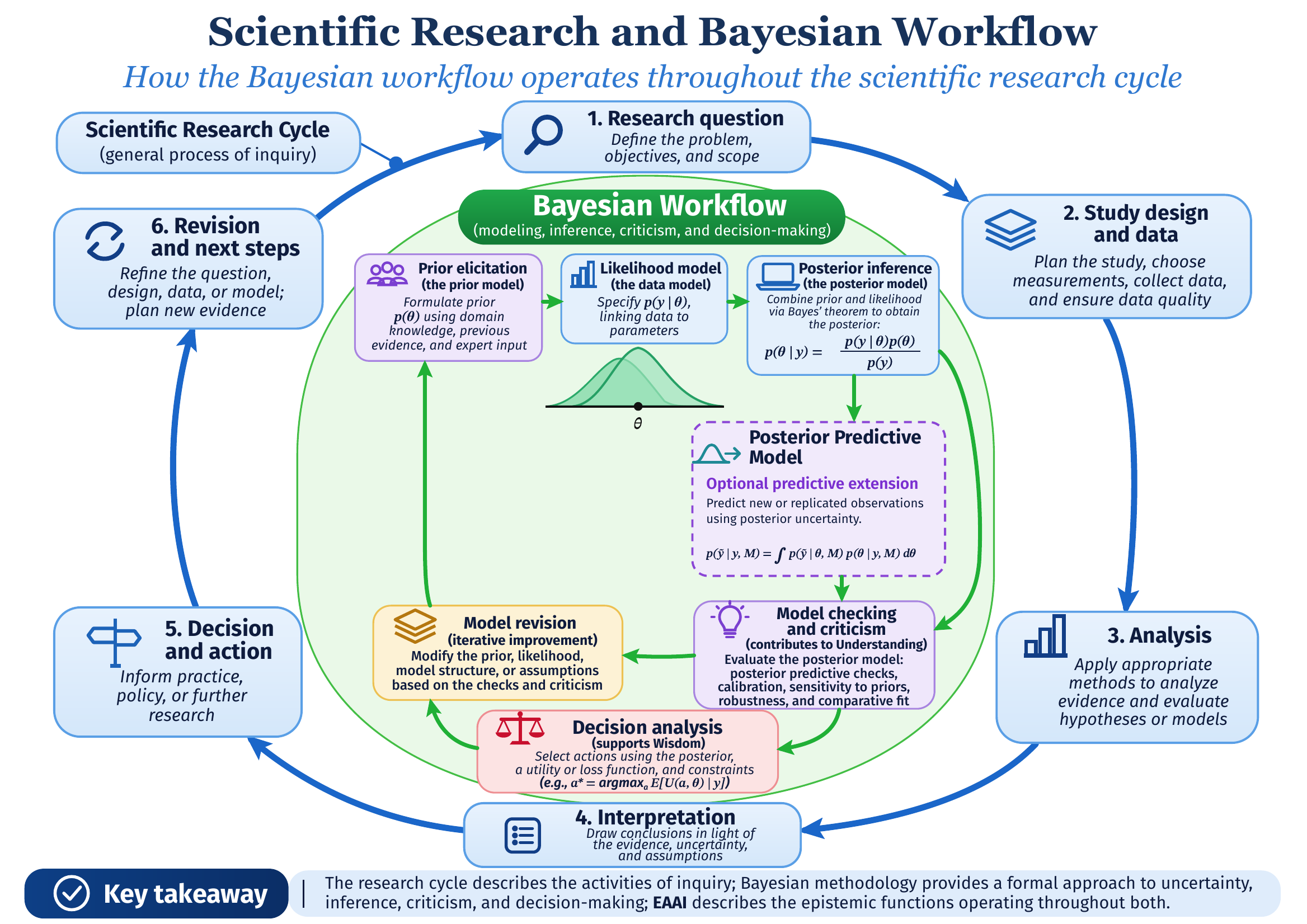}
\caption{Scientific Research and Bayesian Workflow: two interconnected cycles including posterior prediction. The Bayesian practices shown draw on Gelman et al. (2020, 2026).}
\label{fig:workflow}
\end{figure}

Having introduced the general EAAI architecture and its Bayesian interpretation, we can examine how these epistemic functions operate across scientific inquiry. Research is iterative: investigators formulate questions, design studies, gather observations, analyze results, interpret findings, assess limitations and revise their approach. Bayesian methodology supports this process through prior specification, likelihood modeling, posterior inference, posterior prediction, model checking and decisions informed by uncertainty. Figure~\ref{fig:workflow} relates these statistical operations to the broader research cycle without equating either cycle with EAAI itself.

The book Bayesian Workflow (Gelman et al., 2026), by Andrew Gelman and collaborators, provides a substantial account of how applied Bayesian analyses are developed, checked, debugged, compared and improved. Building on earlier work, including a 2020 preprint (Gelman et al., 2020), the book emphasizes iterative modeling, simulation, computational diagnostics, validation and practical case studies. It also includes decision analysis (ch. 20), and Bayesian Data Analysis likewise covers decision theory (Gelman et al., 2013, ch. 9). These contributions describe statistical and computational practice and they should be acknowledged on their own terms.

Posterior inference also enables prediction. The posterior predictive distribution in Equation~\eqref{eq:predictive} integrates uncertainty over model parameters to describe new or replicated observations. Predictions drawn from it support posterior predictive checking and, when combined with goals, utilities and constraints, decision analysis (Gelman et al., 2013, 2026).

Figure~\ref{fig:workflow} connects the scientific research cycle with Bayesian practices that can operate within it. The description of those practices draws on the work of Gelman and collaborators. EAAI supplies a separate interpretive perspective: Data, Information, Knowledge, Understanding and Wisdom can recur throughout the activities depicted, including modeling, prediction, evaluation and action.

The research cycle describes the activities of inquiry. Bayesian methodology describes a formal approach to uncertainty, inference, prediction, criticism and decision-making. EAAI describes the epistemic functions operating throughout both, including their connection to decision-making. The functions can recur, interact and be approached in different orders, and they do not prescribe a one-way pipeline.

\section{Five Kinds of Prior Context—Not Five New Statistical Priors}\label{sec:prior-context}
Here, “prior” is extended in an explicitly epistemic sense: the evidence, interpretations, models, lessons and judgments available before the current inquiry or analysis begins. The five categories in Table~\ref{tab:prior-context} are proposed distinctions in prior context, not five established types of Bayesian prior probability distribution. A mathematical prior distribution is one possible way to encode some of that context. Prior context can also shape the choice of model and likelihood, how the model is evaluated and which actions are considered.

\begin{table}[tbp]
\centering
\caption{Five kinds of prior context in EAAI: proposed distinctions among the evidence, interpretations, models, lessons and judgments available before the current inquiry or analysis begins, not five types of prior probability distribution.}
\label{tab:prior-context}
\small
\renewcommand{\arraystretch}{1.18}
\begin{tabularx}{\linewidth}{@{}>{\raggedright\arraybackslash}p{0.21\linewidth}>{\raggedright\arraybackslash}X@{}}
\toprule
\textbf{Prior context} & \textbf{Description} \\
\midrule
\textbf{Prior Data} & Earlier observations, such as results from previous experiments. Prior Data should remain distinct from the observations under current analysis. Earlier experiments can legitimately inform a prior distribution for a new experiment, but using the same observations both to estimate the prior and to make inferences about them uses the data twice and can overestimate precision. A hierarchical model avoids this by modeling all experiments jointly (Gelman et al., 2013, Section 5.1). \\
\addlinespace
\textbf{Prior Information} & Earlier observations together with their context, measurement conditions and provenance. \\
\addlinespace
\textbf{Prior Knowledge} & Existing patterns, theories, reasons and models. \\
\addlinespace
\textbf{Prior Understanding} & Previously identified failure modes, diagnostic insights and lessons from model criticism. \\
\addlinespace
\textbf{Prior Wisdom} & Existing judgments about goals, values, acceptable risks and consequences. \\
\bottomrule
\end{tabularx}
\end{table}

Prior Information in this sense can concern observations as well as parameters. It is broader than the one definition of prior information found in the examined books, Berger's (1985) ``information about $\theta$ arising from sources other than the statistical investigation'', which concerns the parameter $\theta$ only (Appendix~\ref{app:review}).

These categories can inform inquiry before new observations are collected. Previous model criticism can motivate new checks, existing value judgments can influence the choice of measurements and a decision can expose the need for more evidence. The same recursive relationships will now be examined across three practical AI applications: quantitative finance, actuarial health insurance and agentic reimbursement-claims processing.

\section{EAAI in Practice: Three Worked Examples}\label{sec:examples}
\begin{figure}[tbp]
\centering\includegraphics[width=\linewidth,height=.63\textheight,keepaspectratio]{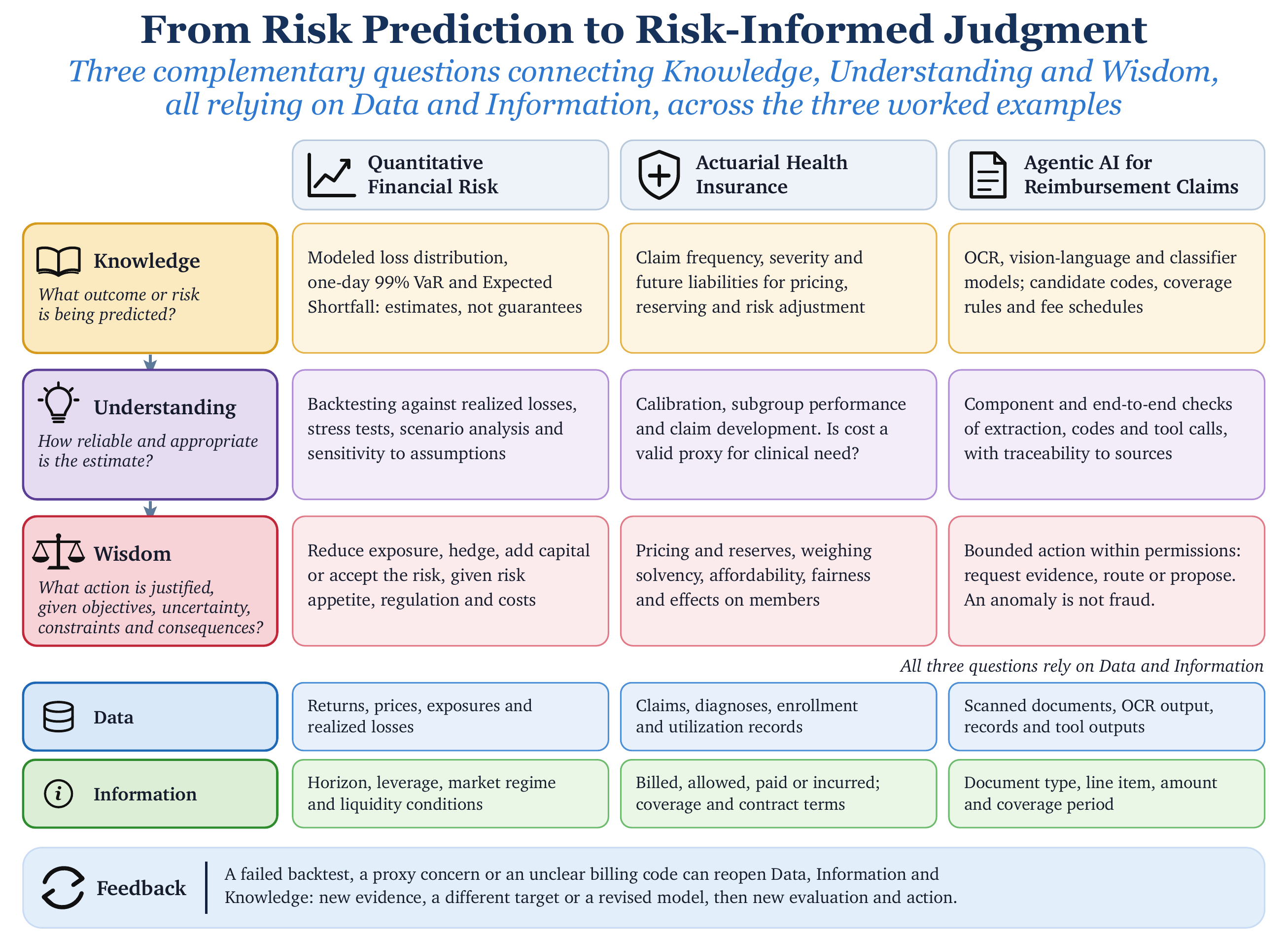}
\caption{From Risk Prediction to Risk-Informed Judgment: three complementary questions connecting Knowledge, Understanding and Wisdom. All three depend on Data and Information and can prompt feedback across the architecture.}
\label{fig:risk}
\end{figure}

The three worked examples apply EAAI to quantitative financial risk, actuarial health insurance and agentic AI for medical reimbursement claims. Figure~\ref{fig:risk} gives an overview of all three: for each example, it shows the risk being predicted (Knowledge), the evaluation of that estimate (Understanding) and the justified action (Wisdom), all relying on Data and Information.

\subsection{Worked Example I: Quantitative Financial Risk}
Consider a financial institution using statistical or machine learning models to assess potential portfolio losses. The exercise demonstrates the distinction between estimating risk, critically evaluating the estimate and deciding how to manage the risk.

\subsubsection{Data: What Has Been Observed?}
Historical returns, asset prices, trading volumes, volatility measures, credit spreads, liquidity measures, portfolio exposures and realized losses provide observations for analysis. Their quality and representativeness matter: an available history may omit severe market conditions, contain inconsistent measurements or describe a portfolio unlike the one presently held.

\subsubsection{Information: What Do the Observations Mean?}
The significance of a portfolio loss depends on the time horizon, portfolio composition, leverage, valuation conventions, market regime and liquidity conditions. Information consists not merely of the numerical records but of their interpretation in this context. Correlations estimated during ordinary markets, for example, may not have the same relevance during market stress.

\subsubsection{Knowledge: What Risk Does the Model Predict?}
A fitted risk model might estimate a distribution of future portfolio losses, a probability of exceeding a loss threshold, Value at Risk (VaR) or Expected Shortfall (ES). Suppose, illustratively, that a model estimates a one-day 99\% VaR of \$2 million and a 99\% ES of \$3.1 million. The former is the modeled 99th percentile of one-day losses. The latter is the modeled mean loss in the worst 1\% of outcomes, where defined. Both estimates depend on assumptions and are not guarantees or upper bounds on realized loss.

\subsubsection{Understanding: How Reliable Are the Estimates?}
Understanding requires examining model specification, historical coverage, sensitivity to assumptions, calibration and behavior during stress. Backtesting compares forecasts with realized outcomes, while stress testing and scenario analysis assess exposures that historical averages may obscure. VaR does not describe the severity of losses beyond its stated percentile. ES addresses their modeled mean severity but is likewise exposed to estimation error. If validation reveals systematic underestimation of stress-period losses, the institution may need to revise assumptions, obtain different evidence or change the model.

\subsubsection{Wisdom: What Should Be Done About the Risk?}
A decision about reducing exposure, hedging, increasing capital buffers or accepting risk cannot be derived from VaR or ES alone. Risk appetite, investment objectives, available capital, liquidity, regulation, hedging costs and possible consequences all influence which action is justified. Two institutions might reasonably take different actions based on similar estimates because their circumstances and constraints differ.

\subsubsection{Recursive Interaction: When Risk Assessment Changes Inquiry}
A failed backtest can initiate new data collection or model revision. A revised risk appetite can shift the modeling objective toward tail losses or liquidity risk. Unexpected losses can trigger reinterpretation of previously observed market conditions. Thus, Understanding may redirect Knowledge and Data, while Wisdom may shape which risks are modeled and how they are evaluated. EAAI clarifies these roles without replacing established financial risk-management methods.

\subsection{Worked Example II: Actuarial Risk Modeling in Health Insurance — From Predictive Accuracy to Responsible Decisions}
Consider a health insurer using actuarial and machine learning models to forecast claims and support pricing, reserving, risk adjustment and financial planning. Actuarial analysis may estimate claim frequency (how often claims occur), severity (their cost when they occur) and aggregate expected expenditure. Although these models can be highly useful, predicting claims costs is not identical to measuring clinical need or determining a fair and appropriate decision. EAAI makes the differences explicit.

\subsubsection{Data: What Has Been Recorded?}
Historical claims, paid and incurred amounts, diagnoses, treatment records, enrollment periods, covered benefits, exposures and healthcare utilization provide observations for actuarial analysis. Claims reported but not yet settled and incomplete reporting periods can complicate estimation. Recorded costs also reflect prices, access, benefit design and administrative processes, not only morbidity. Data therefore require scrutiny before being used to infer either future liabilities or health needs.

\subsubsection{Information: What Do the Records Represent?}
The same claim amount has different implications depending on the coverage period, population mix, contract terms, healthcare prices, cost-sharing arrangements, provider network and whether the amount is billed, allowed, paid or incurred. Actuaries must establish consistent definitions of exposure, claims frequency and severity, development periods and the forecasting horizon. These contextual distinctions turn raw claims records into information relevant to a specified actuarial question. In particular, expenditure is not a direct measure of underlying disease burden or unmet care needs.

\subsubsection{Knowledge: What Does the Model Estimate?}
Models may estimate claim counts, conditional claim severity, total costs per member, morbidity risk or the distribution of future liabilities. Predictions can inform premiums, risk-adjustment calculations, reserves and capital planning, subject to applicable methods and rules. A forecast of next-year costs is Knowledge about the modeled expenditure outcome insofar as its assumptions and evidence warrant that interpretation. It does not automatically establish Knowledge about clinical need, nor does one modeling approach serve all actuarial purposes.

\subsubsection{Understanding: Is the Model Reliable and Is the Target Appropriate?}
Evaluation should examine out-of-sample prediction, calibration, uncertainty, claim development, sensitivity to inflation and utilization trends, changes in coverage and population composition, tail behavior and subgroup performance. A model may predict spending adequately yet be inappropriate when spending is treated as a proxy for medical need. Obermeyer et al. (2019) examined a population health management algorithm, not an insurance pricing model, and found that using healthcare cost as a proxy for health need introduced substantial racial bias. The finding illustrates a general epistemic risk: strong prediction of a recorded target does not establish that the target represents the intended construct.

\subsubsection{Wisdom: How Should Actuarial Estimates Inform Decisions?}
Actuarial projections can inform pricing, reserving, risk adjustment and allocation of financial resources, but they do not uniquely determine those decisions. Decision-makers must consider solvency, affordability, regulatory requirements, uncertainty, distributional effects and potential consequences for members. If the purpose is identifying people who would benefit from additional care, predicted expenditure alone may be an inappropriate allocation criterion. A justified decision requires the target, its validation and the objectives and constraints of its use to be examined together.

\subsubsection{Recursive Interaction: When Actuarial Assessment Changes the Inquiry}
The process may begin with a reserve adequacy concern, a pricing constraint, a population health objective or unexpected experience rather than with new observations. A discrepancy between predicted and realized costs may trigger renewed data checks, changes to exposure definitions or model recalibration. A fairness or clinical-validity concern may lead to a different target and additional evidence. EAAI therefore distinguishes actuarial prediction, evaluation and responsible judgment while showing how each can redirect the others. The example complements, rather than replaces, established actuarial standards and health insurance regulation.

\subsection{Worked Example III: Agentic AI for Medical Reimbursement Claims — From Documents to Risk-Informed Action}
Consider a health insurer deploying an agentic AI system to assist with the review of medical reimbursement claims. Unlike a standalone classifier, the system coordinates several specialized capabilities: document classification, optical character recognition (OCR), extraction of medical and billing information, terminology and code mapping, coverage and reimbursement checks, member and provider history review and detection of potential fraud, waste or abuse. It may request further evidence, reconcile conflicting information and prepare a supported recommendation for an authorized reviewer. Its role is to assist claims assessment within configured permissions, not to declare fraud or independently make consequential adverse determinations.

This example is especially revealing for EAAI because the system is not performing one prediction. It executes a chain of dependent tasks. An OCR error may change the interpretation of a medical procedure, which may lead to an incorrect code, reimbursement comparison or anomaly flag. The epistemic quality of the final outcome depends on the reliability of the intermediate steps and on the controls governing subsequent actions.

\subsubsection{Data: What Documents, Records and Outputs Are Available?}
A reimbursement case may contain scanned invoices, medical reports, prescriptions, procedure descriptions, receipts and claim forms, alongside structured enrollment, claims and provider records. Computer vision classifies incoming documents, while OCR or document-understanding models extract text, amounts, dates, names and other fields. The original document image, extracted text, classification confidence and provenance should remain distinguishable. OCR output is a derived observation that can contain errors, not an unquestionable transcription of the source.

Data also include outputs produced during the workflow: candidate document categories, extracted line items, proposed codes, database query results and rule-check outcomes. Their origin and version matter because information generated by one model may subsequently become input to another.

\subsubsection{Information: What Do These Records Mean for the Claim?}
The agent must establish whether a document is an invoice, a referral, a clinical report or another source; whether extracted dates fall within the relevant coverage period; which procedure each line item describes; and whether the patient, provider and claim refer to the same event. A medical prescription, for example, documents a clinician's order, but does not by itself establish that a medication was dispensed, administered or eligible for reimbursement. The meaning of a monetary amount depends on its currency, units, contract, service setting and whether it represents a billed charge, an allowed amount or a reimbursable benefit.

A missing supporting report is not evidence that a procedure did not occur, and two similar claims are not automatically duplicates. The system needs to preserve distinctions among the original source, its machine interpretation and unresolved discrepancies. If documents conflict, the agent can retrieve another authorized source or request clarification rather than silently resolving the contradiction.

\subsubsection{Knowledge: Which Models, Code Systems and Rules Support the Assessment?}
The agent may use a document classifier, language or vision-language models for extraction, a terminology-mapping component and specialized services that recommend standardized medical or billing codes. In European settings, diagnosis coding may draw on the World Health Organization's International Classification of Diseases (ICD) or national adaptations. Procedure, billing and reimbursement coding is not uniform across Europe and depends on the country, healthcare setting, payer and applicable code-set version. The system should present candidate codes with supporting document spans and reasons, especially when descriptions are ambiguous or insufficient.

Additional components may check benefit eligibility, prior authorizations, exclusions, contracted fee schedules, coding compatibility and billed versus allowed amounts. Another model may estimate the likelihood that a claim warrants further investigation using patterns such as unusual charges, repeated services or inconsistent provider behavior. These distinct resources belong to Knowledge as models, rules and learned representations, but they answer different questions. A medically plausible code does not prove that the service was delivered; a valid code does not establish coverage; an anomalous charge does not prove fraud.

\subsubsection{Understanding: Are the Documents, Codes, Prices and Risk Flags Reliable?}
Critical evaluation must operate at the component and end-to-end levels. The insurer should test document classification, OCR field accuracy, clinical concept extraction, code agreement with qualified coders, policy-rule correctness, fee-schedule matching, predictive calibration and the performance of fraud-risk screening. It must also test whether the agent calls the correct tools, respects access restrictions, handles ambiguous documents and maintains traceability from each conclusion to its source. Conventional OCR engines, such as Tesseract or the text recognizers in PaddleOCR, typically err at the character level or produce visibly garbled text, whereas vision-language models, including OCR systems built on them, can produce fluent words or values absent from the source (He et al., 2025; Gardella et al., 2026). Extracted values should therefore be traced to the document regions they came from.

Suppose a scanned invoice appears to describe a procedure that normally has an allowed reimbursement of \$400, but the submitted amount is \$1,200. The agent should first verify the OCR extraction, the precise procedure and code, service location, units, contract and applicable benefit rules. The difference may reflect a billing error, a different procedure, multiple units, an out-of-network arrangement or an anomaly requiring investigation. Price comparison alone cannot establish wrongdoing.

Likewise, examining a member’s history may reveal a repeated claim or coverage inconsistency, while provider-level patterns may suggest systematic overbilling. Such indicators require contextual assessment and appropriate safeguards against unsupported conclusions and unfair profiling. Validation should distinguish false positives, which may cause unnecessary delays or allegations, from false negatives, which may permit improper payments. A technically successful multi-step workflow can still be epistemically unsound if it propagates one unnoticed extraction error.

\subsubsection{Wisdom: What Action Is Justified and Who May Authorize It?}
The agent may, within explicit authority, assemble an evidence-linked claim summary, recommend a candidate code, identify discrepancies, request missing documentation, route a case to a specialist or propose an amount for review. It should not automatically equate anomaly detection with a finding of fraud. Determinations affecting reimbursement, accusations of misconduct, adverse member outcomes or referral for investigation must follow applicable contractual, professional, organizational and legal procedures, with appropriate human authority, explanation and appeal or correction mechanisms.

The decision involves more than minimizing payment losses. The insurer must consider accuracy, timely reimbursement, legitimate member and provider interests, operational cost, privacy, fairness, and the harms of delaying or incorrectly rejecting valid claims. Wisdom here refers to the quality and justification of decision policy and oversight, not to attributing moral judgment to the software agent.

\subsubsection{Recursive Interaction: A Worked Claim-Level Sequence}
Imagine the document classifier recognizes an invoice and a medical report. OCR extracts a procedure description and a billed amount, but the terminology mapper proposes two plausible billing codes. Instead of selecting one without qualification, the agent checks the clinical report and finds that a crucial procedure detail is missing. It requests additional documentation or escalates to a medical coder. Once clarified, the revised code leads to a new fee-schedule lookup and changes the reimbursement discrepancy. A fraud-screening flag may then disappear, remain open as an unexplained anomaly or be escalated with its supporting evidence. The intended action changes as the evidence and interpretation change.

The workflow can also begin elsewhere. A claim flagged for an unusually high amount may cause the agent to revisit the scanned invoice; an updated reimbursement contract may require all affected claims to be rechecked; a human reviewer may question the predicted code and trigger new extraction or verification. The five EAAI functions therefore operate recursively, rather than forming fixed stages in the same order as the system’s technical pipeline.

\subsubsection{What Agentic AI Adds to the Epistemic Risk}
An error by a standalone classifier remains confined to that prediction. An agent can compound an error across document reading, interpretation, code selection, pricing logic, history queries and tool-mediated actions. Other hazards include retrieving stale policies, following malicious instructions embedded in uploaded documents, exposing sensitive patient data or modifying a claim record without authorization. An appropriate evaluation must therefore examine the entire sequence of evidence, intermediate decisions, tool calls and final actions, not merely an isolated accuracy score.

This example demonstrates the practical role of EAAI: Data identifies the available records and generated outputs; Information establishes their meaning in context; Knowledge provides models, codes and rules; Understanding evaluates reliability, ambiguity and failure; Wisdom governs justified and accountable action. A discrepancy between medical documentation, billing codes and claimed reimbursement amounts does not, by itself, establish fraud. It may reflect an administrative error, incomplete documentation, incorrect coding or other circumstances requiring investigation. The system should identify and explain potential anomalies rather than automatically classify a claim as fraudulent. Specific coding, reimbursement and authorization rules depend on the jurisdiction and insurance product.

\subsection{What the Risk and Agentic AI Examples Demonstrate}
Across quantitative financial risk, actuarial health insurance modeling and agentic reimbursement-claims processing, the examples expose three analytically distinguishable questions: What outcome or risk is being predicted? How reliable and appropriate is the estimate or intermediate conclusion? What action is justified given objectives, uncertainty, constraints and consequences? These questions are associated primarily with Knowledge, Understanding and Wisdom, while relying throughout on Data and Information. Figure~\ref{fig:risk} sets these questions against the three examples.

Taken together, the examples show how EAAI distinguishes evidence, contextual interpretation, models and predictions, critical appraisal and informed judgment. They also demonstrate how these functions recur and affect one another across financial risk management, actuarial analysis and agentic reimbursement-claims processing.
\section{Implications for AI and Machine Learning}
EAAI does not introduce new methods. It gives established practices a common vocabulary and a place on a single map, and the map makes their gaps and dependencies visible. This section draws the consequences for how AI systems are built, evaluated and governed.

\textbf{Evaluation by Function.} A model's predictive performance answers only one of three questions. Accuracy, loss and error on held-out data measure how well a model predicts under conditions like those of its test data, the quality of the Knowledge it encodes. They do not establish Understanding, which requires evidence about reliability beyond those conditions: under distribution shift, in calibration, across subgroups and in failure modes, since predictors with equivalent test performance can behave very differently in deployment (D'Amour et al., 2022). Nor do they establish Wisdom, which concerns the consequences of the decisions a model informs. A successful prediction does not by itself establish an adequate explanation or an appropriate proxy for the decision target. Reporting evidence separately for each function would show which questions an evaluation has addressed and which remain open. Relevant practices include out-of-sample validation, calibration, ablations, sensitivity analysis, stress testing, distribution-shift assessment, residual and failure analysis and causal investigation when causal claims are made (Gelman et al., 2026; Bishop, 2006; Murphy, 2023).

\textbf{Functions as Roles.} The same component can serve different functions within one system. An OCR or vision-language model belongs to Knowledge: a learned representation that maps images to text. Its output becomes Data for the next component, which must interpret it before relying on it. In agentic systems, where one model's output is another's input, an error that enters as a derived observation can pass through interpretation, coding, pricing and action without ever being recognized as an error. Each intermediate output therefore needs its provenance recorded and should be treated as an observation open to challenge, not as an established fact.

\textbf{Independent Evidence.} Using the same observations both to estimate a prior and to make inferences about them overstates precision (Section~\ref{sec:prior-context}). Machine learning has the same structure: data that have guided model selection, tuning or prompt design no longer provide an independent test of the resulting system. Held-out test sets are one familiar application of a general principle: Prior Data must remain distinct from the observations used to evaluate the models built from them.

\textbf{Generation and Appraisal.} Generative models produce outputs that read as conclusions. When the same system both produces an answer and is trusted to assess it, Knowledge and Understanding collapse into one step, a risk closely related to the condition Quattrociocchi et al. (2025) describe, in which “linguistic plausibility substitutes for epistemic evaluation.” EAAI implies that appraisal should remain separate from generation: checks against sources, independent validation and explicit uncertainty, rather than reliance on the fluency or stated confidence of the generated text.

\textbf{Uncertainty Across Functions.} Uncertainty quantification links three functions. A predictive distribution or interval is produced as Knowledge, appraised for calibration in Understanding and given its value in Wisdom, through the decision it informs. Distinguishing epistemic from aleatoric uncertainty (Hüllermeier and Waegeman, 2021) indicates when more data or a better model can reduce uncertainty, and when a decision must instead accommodate variability that will remain.

\textbf{Placement of Oversight.} Human review applied uniformly to every step is costly and can become routine. EAAI locates its greatest value at two points: Understanding, where evidence about reliability determines whether an output may be relied on at all, and Wisdom, where an output becomes a decision with consequences for people. In the reimbursement-claims example, the agent assembles evidence and proposes actions, and the authority to deny a claim or to refer it for investigation remains with accountable people.

\textbf{Shared Vocabulary.} Bayesian methodology already provides explicit tools for prior context, belief updating, prediction, model criticism and decision analysis, yet its terminology is often unfamiliar outside statistics. Mapping these tools to five intuitive functions makes their purpose legible to engineers, domain experts and decision makers who do not use Bayesian language. The same map applies to systems that are not Bayesian: wherever a model is trained, evaluated and used to act, the five functions are present, whether or not they are named, and the map exposes the ones a pipeline leaves implicit.

Across AI development and use, an error in interpretation, a failed validation check or a change in decision requirements can prompt renewed inquiry. EAAI makes these points of revision and their epistemic roles explicit.

\section{Scope and Limitations of the Proposal}
The scope of EAAI is conceptual and analytical: its usefulness lies in the clarity of its distinctions, the coherence of its relationships and its ability to illuminate epistemic questions across different AI methodologies and applications. Its five functions are not claims that each AI system independently possesses human-like Understanding or Wisdom. These functions can also characterize the practices of people and institutions that develop, evaluate and use AI.

The targeted review supports only a bounded observation about the examined books: a systematic distinction among these five epistemic functions was not identified in the passages reviewed. It does not show that such a distinction is absent from the wider literature, nor that EAAI is the first framework to examine these categories. Frické (2009), for example, criticized the DIKW hierarchy itself, and Aven (2013) linked its elements to risk assessment. Section~\ref{sec:related} compares EAAI with these and other related work.

\section{Related Work and Critical Comparison}\label{sec:related}
EAAI draws on several bodies of work. This section relates it to interpretations and critiques of the DIKW hierarchy, philosophical work on understanding, Bayesian workflow and model criticism, cognitive architectures, AI evaluation practice and risk-governance frameworks. For each, it identifies the ideas EAAI shares with that work and the ideas EAAI adds. The comparison is selective rather than exhaustive. Table~\ref{tab:related} summarizes it.

\subsection{DIKW Interpretations and Critiques}
Ackoff (1989) presented Data, Information, Knowledge, Understanding and Wisdom as a hierarchy, and EAAI takes its five categories from that formulation. Later reviews documented how unsettled the categories remain. Rowley (2007) found that each element is typically defined in terms of the one below it, with less consensus on the processes that transform one into the next and limited reference to wisdom. Zins (2007) documented 130 definitions of data, information and knowledge formulated by 45 scholars, and Baškarada and Koronios (2013) identified a range of definitional ambiguities from a practical perspective.

Other authors questioned the direction of the hierarchy itself. Tuomi (1999) argued that data emerge only after there is information, and information only after there is knowledge. Frické (2009) argued that the hierarchy is unsound and methodologically undesirable, and identified operationalism and inductivism as its philosophical backdrop. Aven (2013) linked the DIKW elements to roles in risk assessment, from the input data to the decision maker's use of the results, and treated understanding as part of knowledge rather than as a separate element. More recent work connects these critiques to AI. Calzati (2025) proposes an ecosystemic rearrangement of data, information and knowledge with agential AI in view. Cristofaro and Bañón-Gomis (2026) reinterpret DIKW “not as a linear information hierarchy” and map AI-assisted managerial decisions to four recursive phases: frame, evaluate, commit and enact.

EAAI shares the rejection of a one-way derivation from data upward with Tuomi, with Frické and with Cristofaro and Bañón-Gomis, and a role-based reading of the elements with Aven. It differs in three respects. It keeps all five categories, including Understanding, but defines each as an epistemic function rather than as a level or a type of content. It gives Understanding the specific role of critical evaluation and Wisdom the role of justifying action. And it applies these functions to AI and machine learning, through a Bayesian interpretation and through predictive and agentic applications.

\subsection{Understanding as a Distinct Epistemic Achievement}
De Regt (2017) observes that most philosophers regarded understanding “as merely a subjective byproduct of objective explanations” and develops a theory that puts scientific understanding center stage. Grimm (2006) defends the opposing position that understanding is a species of knowledge. EAAI does not depend on settling this debate. Its distinction between Knowledge and Understanding is functional: producing a model or a prediction differs from critically appraising why it succeeds or fails, whether or not that appraisal is classified as a form of knowledge.

\subsection{Bayesian Workflow and Model Criticism}
Bayesian practice already separates fitting a model from checking it. Gelman and Shalizi (2013) argue that the most successful forms of Bayesian statistics accord with sophisticated forms of hypothetico-deductivism, and that model checking and model revision “fall outside the scope of Bayesian confirmation theory.” Blei (2014) describes the same iteration: “We build a model, use it to analyze data, assess how it succeeds and fails, revise it, and repeat.” Gelman et al. (2020, 2026) develop it into a workflow for applied Bayesian analysis. EAAI agrees with this iterative view and places it in a wider frame. It adds the distinction between Data and Information before modeling begins, the role of Wisdom in decisions after it and the five kinds of prior context in Table~\ref{tab:prior-context}.

\subsection{Cognitive Architectures}
In AI, “architecture” often denotes a computational theory of cognition. Soar is “both a software system for agent development and a theory of what computational structures are necessary to support human-level agents” (Laird, 2012). ACT-R consists of multiple modules and explains how they “are integrated to produce coherent cognition” (Anderson et al., 2004), and Kotseruba and Tsotsos (2020) survey 84 such architectures. EAAI uses the term in a different sense. It does not specify the modules or mechanisms of an agent. It distinguishes epistemic functions that people, institutions and AI systems can perform, and examines how those functions depend on one another.

\subsection{AI Evaluation, Documentation and Uncertainty}
Several strands of machine learning research address the functions EAAI calls Data, Information and Understanding. Datasheets document a dataset's motivation, composition, collection process and recommended uses (Gebru et al., 2021), supporting the provenance and context that turn Data into Information. Model cards report the evaluation of trained models across a range of conditions, such as demographic groups (Mitchell et al., 2019), and internal algorithmic auditing extends evaluation across the development lifecycle (Raji et al., 2020). D'Amour et al. (2022) show that predictors with equivalently strong test performance “can behave very differently in deployment domains,” which supports separating a model's predictive performance, a matter of Knowledge, from the appraisal of its reliability, a matter of Understanding. Hüllermeier and Waegeman (2021) emphasize distinguishing aleatoric from epistemic uncertainty, a distinction central to uncertainty quantification within that appraisal. For language models, Quattrociocchi et al. (2025) describe a condition in which “linguistic plausibility substitutes for epistemic evaluation,” a failure related to the vision-language extraction risk in the reimbursement-claims example. EAAI does not replace these practices. It locates each within the function it serves and shows how its results can feed back into data collection, modeling and decisions.

\subsection{Risk-Governance Frameworks}
The NIST AI Risk Management Framework organizes AI risk management into four functions, Govern, Map, Measure and Manage. Governance is “designed to be a cross-cutting function,” and the framework's actions “do not constitute a checklist, nor are they necessarily an ordered set of steps” (NIST, 2023). In the European Union, the AI Act classifies AI systems intended for risk assessment and pricing in relation to natural persons in life and health insurance as high-risk (Regulation (EU) 2024/1689, Annex III, point 5(c)). Such systems require a risk management system run as “a continuous iterative process” throughout their lifecycle (Article 9) and human oversight that enables people “to decide, in any particular situation, not to use the high-risk AI system or to otherwise disregard, override or reverse the output of the high-risk AI system” (Article 14(4)(d)). For systems classified as high-risk under Annex III, these obligations apply from 2 December 2027 (Regulation (EU) 2026/1744). For uses outside the high-risk category, such as claims management and fraud detection, the Opinion on AI governance and risk management of the European Insurance and Occupational Pensions Authority sets supervisory expectations, including that “the outcomes of AI systems can be meaningfully explained” (EIOPA, 2025). In banking, EU law defines model risk as “the risk of loss resulting from decisions that are principally based on the output of internal models, due to errors in the design, development, parameter estimation, implementation, use or monitoring of such models” (Regulation (EU) No 575/2013, Article 4(1)(52b), inserted by Regulation (EU) 2024/1623), and the ECB guide to internal models requires that internal validation “effectively challenge the modelling decisions taken regarding ML-based internal models” (ECB, 2026). These frameworks organize risk-management activities, obligations and responsibilities. EAAI is complementary: it distinguishes the epistemic roles of the observations, interpretations, models, evaluations and judgments on which those activities rely, such as the evidence a human overseer needs before overriding an output.

\begin{table}[tbp]
\centering
\caption{EAAI compared with related work: the ideas each strand shares with EAAI and the ideas EAAI adds. Selective comparison based on the sources discussed in Section~\ref{sec:related}.}
\label{tab:related}
\small
\renewcommand{\arraystretch}{1.18}
\begin{tabularx}{\linewidth}{@{}>{\raggedright\arraybackslash}p{0.25\linewidth}>{\raggedright\arraybackslash}X>{\raggedright\arraybackslash}X@{}}
\toprule
\textbf{Strand} & \textbf{Shared with EAAI} & \textbf{Added by EAAI} \\
\midrule
\textbf{DIKW interpretations and critiques} \newline Ackoff (1989); Rowley (2007); Tuomi (1999); Frické (2009); Aven (2013); Cristofaro and Bañón-Gomis (2026) & The five categories; rejection of a one-way derivation; recursive phases; role-based readings & Categories defined as epistemic functions; Understanding as critical evaluation and Wisdom as justified action; application to AI and machine learning \\
\addlinespace
\textbf{Philosophy of understanding} \newline de Regt (2017); Grimm (2006) & Understanding as an epistemic achievement in its own right (de Regt) & A functional distinction that holds whether or not understanding is a species of knowledge \\
\addlinespace
\textbf{Bayesian workflow and model criticism} \newline Gelman and Shalizi (2013); Blei (2014); Gelman et al. (2020, 2026) & Iteration; model checking separated from model fitting & Adds the distinction between Data and Information before modeling, the role of Wisdom in decisions and the five kinds of prior context \\
\addlinespace
\textbf{Cognitive architectures} \newline Laird (2012); Anderson et al. (2004); Kotseruba and Tsotsos (2020) & Distinct components that are integrated & Epistemic functions of people, institutions and systems, not mechanisms of an agent \\
\addlinespace
\textbf{AI evaluation, documentation and uncertainty} \newline Gebru et al. (2021); Mitchell et al. (2019); Raji et al. (2020); D'Amour et al. (2022); Hüllermeier and Waegeman (2021); Quattrociocchi et al. (2025) & Documenting where data come from (datasheets); evaluating models beyond a single test score (model cards, audits, underspecification); quantifying uncertainty & Assigns each practice to the function it serves: datasheets to Data and Information; evaluation, auditing and uncertainty quantification to Understanding. Shows how a failed check leads back to new data, a revised model or a different decision \\
\addlinespace
\textbf{Risk-governance frameworks} \newline NIST (2023); Regulation (EU) 2024/1689; EIOPA (2025); Regulation (EU) 2024/1623; ECB (2026) & Functions rather than ordered steps; lifecycle risk management; human oversight; challenge of models & Distinguishes the epistemic roles of the observations, models, evaluations and judgments that risk management and human oversight rely on \\
\bottomrule
\end{tabularx}
\end{table}

\subsection{Precedents and Contribution}
Within the sources reviewed here, each element of EAAI has a precedent: non-linear readings of DIKW, recursive decision phases, the separation of model checking from model fitting and the separation of evaluation from decision in risk governance. EAAI contributes their combination: a single vocabulary of five functions defined by epistemic role, explicit recursive relationships among them, a Bayesian interpretation, five kinds of prior context and application to predictive and agentic AI. Whether this combination has precedents beyond the sources examined would require a more systematic review.

\section{Conclusion}
The Data-to-Wisdom Epistemic Architecture for AI (EAAI) treats Data, Information, Knowledge, Understanding and Wisdom as five distinct but recursively interacting epistemic functions: observation, contextual interpretation, modeling and prediction, critical evaluation and informed judgment. It applies to AI and machine learning generally, Bayesian or not, and to the people and institutions that build, evaluate and use AI systems. Restoring Understanding, the level that later versions of the DIKW hierarchy omit, gives critical evaluation a place of its own between the models that produce outputs and the judgments that act on them.

Bayesian methodology offers the most explicit methodological reading of these functions. Relating the likelihood, prior and posterior distributions, posterior prediction, model checking and decision analysis to the five functions makes the purpose of Bayesian tools accessible beyond statistics. The five kinds of prior context show that resources available before an analysis can shape the model, the likelihood, the prior, evaluation and action, and the distinctions among prior information, prior knowledge and evidence clarify terminology that the examined literature leaves ambiguous.

The worked examples in quantitative finance, actuarial health insurance and agentic reimbursement-claims processing show the same pattern: predicting a risk, appraising the reliability of that prediction and justifying an action are different questions, each relying on Data and Information. Across these settings, EAAI helps distinguish the observations of AI systems, the meaning of their outputs, the assessment of their reliability and the justification of decisions. For AI and machine learning, this implies evaluating each function on its own evidence, treating intermediate model outputs as observations open to challenge, keeping the appraisal of generated outputs separate from their generation and placing human oversight where reliability is judged and action is authorized.

EAAI is grounded in Ackoff's five categories, which suffice for the analyses presented here. The architecture could be extended with further functions, and its relation to prior work would benefit from a more systematic review. As AI systems increasingly combine models, tools and actions, distinguishing the epistemic roles of their components offers a way to examine their dependencies, feedback, limitations and consequences.

\section*{References}
\begingroup
\small
\raggedright
\setlength{\parindent}{0pt}
\setlength{\parskip}{0.35em}
\hangindent=1.3em\hangafter=1
Ackoff, R. L. (1989). From Data to Wisdom. Journal of Applied Systems Analysis, 16, 3–9.

\hangindent=1.3em\hangafter=1
Anderson, J. R., Bothell, D., Byrne, M. D., Douglass, S., Lebiere, C., and Qin, Y. (2004). An integrated theory of the mind. Psychological Review, 111(4), 1036–1060. \url{https://doi.org/10.1037/0033-295X.111.4.1036}

\hangindent=1.3em\hangafter=1
Aven, T. (2013). A conceptual framework for linking risk and the elements of the data–information–knowledge–wisdom (DIKW) hierarchy. Reliability Engineering \& System Safety, 111, 30–36. \url{https://doi.org/10.1016/j.ress.2012.09.014}

\hangindent=1.3em\hangafter=1
Barber, D. (2012). Bayesian Reasoning and Machine Learning. Cambridge University Press. \url{https://doi.org/10.1017/CBO9780511804779}

\hangindent=1.3em\hangafter=1
Baškarada, S., and Koronios, A. (2013). Data, information, knowledge, wisdom (DIKW): A semiotic theoretical and empirical exploration of the hierarchy and its quality dimension. Australasian Journal of Information Systems, 18(1), 5–24. \url{https://doi.org/10.3127/ajis.v18i1.748}

\hangindent=1.3em\hangafter=1
Berger, J. O. (1985). Statistical Decision Theory and Bayesian Analysis (2nd ed.). Springer. \url{https://doi.org/10.1007/978-1-4757-4286-2}

\hangindent=1.3em\hangafter=1
Bishop, C. M. (2006). Pattern Recognition and Machine Learning. Springer.

\hangindent=1.3em\hangafter=1
Blei, D. M. (2014). Build, compute, critique, repeat: Data analysis with latent variable models. Annual Review of Statistics and Its Application, 1, 203–232. \url{https://doi.org/10.1146/annurev-statistics-022513-115657}

\hangindent=1.3em\hangafter=1
Calzati, S. (2025). An ecosystemic view on information, data, and knowledge: Insights on agential AI and relational ethics. AI and Ethics, 5(4), 3763–3776. \url{https://doi.org/10.1007/s43681-025-00665-0}

\hangindent=1.3em\hangafter=1
Cristofaro, M., and Bañón-Gomis, A. J. (2026). Dancing with the algorithm: A framework to navigate knowledge and autonomy in AI-assisted managerial decisions. Journal of Knowledge Management, 30(11), 1–31. \url{https://doi.org/10.1108/JKM-06-2025-0870}

\hangindent=1.3em\hangafter=1
D'Amour, A., Heller, K., Moldovan, D., et al. (2022). Underspecification presents challenges for credibility in modern machine learning. Journal of Machine Learning Research, 23(226), 1–61. \url{https://jmlr.org/papers/v23/20-1335.html}

\hangindent=1.3em\hangafter=1
Eliot, T. S. (1934). The Rock: A Pageant Play. Faber \& Faber. [Quotation from “Choruses from The Rock.”]

\hangindent=1.3em\hangafter=1
European Central Bank (ECB). (2026). ECB guide to internal models (June 2026). \url{https://www.bankingsupervision.europa.eu/ecb/pub/pdf/ssm.supervisory_guide_egim_202606.en.pdf}

\hangindent=1.3em\hangafter=1
European Insurance and Occupational Pensions Authority (EIOPA). (2025). Opinion on artificial intelligence governance and risk management (EIOPA-BoS-25-360). \url{https://www.eiopa.europa.eu/publications/opinion-artificial-intelligence-governance-and-risk-management_en}

\hangindent=1.3em\hangafter=1
Frické, M. (2009). The knowledge pyramid: A critique of the DIKW hierarchy. Journal of Information Science, 35(2), 131–142. \url{https://doi.org/10.1177/0165551508094050}

\hangindent=1.3em\hangafter=1
Frické, M. (2019). The knowledge pyramid: The DIKW hierarchy. Knowledge Organization, 46(1), 33–46. \url{https://doi.org/10.5771/0943-7444-2019-1-33}. Also available in B. Hjørland and C. Gnoli (Eds.), ISKO Encyclopedia of Knowledge Organization. \url{https://www.isko.org/cyclo/dikw}

\hangindent=1.3em\hangafter=1
Gardella, M., Mariño, C., Belzarena, D., Ramírez, I., Randall, G., and Morel, J.-M. (2026). When Low CER is Not Enough: An Analysis of Hallucinations in Vision-Language OCR Systems on Historical Uruguayan Documents. ADAPDA 2026 Workshop, ICDAR 2026. arXiv:2607.24077. \url{https://arxiv.org/abs/2607.24077}

\hangindent=1.3em\hangafter=1
Garnett, R. (2023). Bayesian Optimization. Cambridge University Press.

\hangindent=1.3em\hangafter=1
Gebru, T., Morgenstern, J., Vecchione, B., Vaughan, J. W., Wallach, H., Daumé III, H., and Crawford, K. (2021). Datasheets for datasets. Communications of the ACM, 64(12), 86–92. \url{https://doi.org/10.1145/3458723}

\hangindent=1.3em\hangafter=1
Gelman, A., Carlin, J. B., Stern, H. S., Dunson, D. B., Vehtari, A., and Rubin, D. B. (2013). Bayesian Data Analysis (3rd ed.). CRC Press.

\hangindent=1.3em\hangafter=1
Gelman, A., and Shalizi, C. R. (2013). Philosophy and the practice of Bayesian statistics. British Journal of Mathematical and Statistical Psychology, 66(1), 8–38. \url{https://doi.org/10.1111/j.2044-8317.2011.02037.x}

\hangindent=1.3em\hangafter=1
Gelman, A., et al. (2020). Bayesian workflow [preprint]. arXiv:2011.01808. \url{https://arxiv.org/abs/2011.01808}

\hangindent=1.3em\hangafter=1
Gelman, A., Vehtari, A., McElreath, R., et al. (2026). Bayesian Workflow. Chapman \& Hall/CRC. \url{https://doi.org/10.1201/9781003044024}

\hangindent=1.3em\hangafter=1
Grimm, S. R. (2006). Is understanding a species of knowledge? The British Journal for the Philosophy of Science, 57(3), 515–535. \url{https://doi.org/10.1093/bjps/axl015}

\hangindent=1.3em\hangafter=1
He, Z., Zhang, C., Wu, Z., Chen, Z., Zhan, Y., Li, Y., Zhang, Z., Wang, X., and Qiu, M. (2025). Seeing is Believing? Mitigating OCR Hallucinations in Multimodal Large Language Models. Advances in Neural Information Processing Systems (NeurIPS 2025). arXiv:2506.20168. \url{https://arxiv.org/abs/2506.20168}

\hangindent=1.3em\hangafter=1
Hüllermeier, E., and Waegeman, W. (2021). Aleatoric and epistemic uncertainty in machine learning: An introduction to concepts and methods. Machine Learning, 110(3), 457–506. \url{https://doi.org/10.1007/s10994-021-05946-3}

\hangindent=1.3em\hangafter=1
Kochenderfer, M. J., Wheeler, T. A., and Wray, K. H. (2022). Algorithms for Decision Making. MIT Press. \url{https://algorithmsbook.com/}

\hangindent=1.3em\hangafter=1
Kotseruba, I., and Tsotsos, J. K. (2020). 40 years of cognitive architectures: Core cognitive abilities and practical applications. Artificial Intelligence Review, 53(1), 17–94. \url{https://doi.org/10.1007/s10462-018-9646-y}

\hangindent=1.3em\hangafter=1
Laird, J. E. (2012). The Soar Cognitive Architecture. MIT Press. \url{https://doi.org/10.7551/mitpress/7688.001.0001}

\hangindent=1.3em\hangafter=1
MacKay, D. J. C. (2003). Information Theory, Inference, and Learning Algorithms. Cambridge University Press.

\hangindent=1.3em\hangafter=1
Mitchell, M., Wu, S., Zaldivar, A., Barnes, P., Vasserman, L., Hutchinson, B., Spitzer, E., Raji, I. D., and Gebru, T. (2019). Model cards for model reporting. In Proceedings of the Conference on Fairness, Accountability, and Transparency (FAT* '19) (pp. 220–229). ACM. \url{https://doi.org/10.1145/3287560.3287596}

\hangindent=1.3em\hangafter=1
Murphy, K. P. (2022). Probabilistic Machine Learning: An Introduction. MIT Press.

\hangindent=1.3em\hangafter=1
Murphy, K. P. (2023). Probabilistic Machine Learning: Advanced Topics. MIT Press.

\hangindent=1.3em\hangafter=1
National Institute of Standards and Technology (NIST). (2023). Artificial Intelligence Risk Management Framework (AI RMF 1.0) (NIST AI 100-1). \url{https://doi.org/10.6028/NIST.AI.100-1}

\hangindent=1.3em\hangafter=1
Obermeyer, Z., Powers, B., Vogeli, C., and Mullainathan, S. (2019). Dissecting racial bias in an algorithm used to manage the health of populations. Science, 366(6464), 447–453. \url{https://doi.org/10.1126/science.aax2342}

\hangindent=1.3em\hangafter=1
Quattrociocchi, W., Capraro, V., and Perc, M. (2025). Epistemological fault lines between human and artificial intelligence [preprint]. arXiv:2512.19466. \url{https://arxiv.org/abs/2512.19466}

\hangindent=1.3em\hangafter=1
Raji, I. D., Smart, A., White, R. N., Mitchell, M., Gebru, T., Hutchinson, B., Smith-Loud, J., Theron, D., and Barnes, P. (2020). Closing the AI accountability gap: Defining an end-to-end framework for internal algorithmic auditing. In Proceedings of the 2020 Conference on Fairness, Accountability, and Transparency (FAT* '20) (pp. 33–44). ACM. \url{https://doi.org/10.1145/3351095.3372873}

\hangindent=1.3em\hangafter=1
Rasmussen, C. E., and Williams, C. K. I. (2006). Gaussian Processes for Machine Learning. MIT Press.

\hangindent=1.3em\hangafter=1
de Regt, H. W. (2017). Understanding Scientific Understanding. Oxford University Press. \url{https://doi.org/10.1093/oso/9780190652913.001.0001}

\hangindent=1.3em\hangafter=1
Regulation (EU) 2024/1623 of the European Parliament and of the Council of 31 May 2024 amending Regulation (EU) No 575/2013 as regards requirements for credit risk, credit valuation adjustment risk, operational risk, market risk and the output floor. Official Journal of the European Union, L, 2024/1623, 19 June 2024. \url{http://data.europa.eu/eli/reg/2024/1623/oj}

\hangindent=1.3em\hangafter=1
Regulation (EU) 2024/1689 of the European Parliament and of the Council of 13 June 2024 laying down harmonised rules on artificial intelligence (Artificial Intelligence Act). Official Journal of the European Union, L, 2024/1689, 12 July 2024. \url{http://data.europa.eu/eli/reg/2024/1689/oj}

\hangindent=1.3em\hangafter=1
Regulation (EU) 2026/1744 of the European Parliament and of the Council of 8 July 2026 amending Regulations (EU) 2024/1689, (EU) 2018/1139 and (EU) 2023/1230 as regards the simplification of the implementation of harmonised rules on artificial intelligence (Digital Omnibus on AI). Official Journal of the European Union, L, 2026/1744, 24 July 2026. \url{http://data.europa.eu/eli/reg/2026/1744/oj}

\hangindent=1.3em\hangafter=1
Rowley, J. (2007). The Wisdom Hierarchy: Representations of the DIKW Hierarchy. Journal of Information Science, 33(2), 163–180.

\hangindent=1.3em\hangafter=1
Tuomi, I. (1999). Data is more than knowledge: Implications of the reversed knowledge hierarchy for knowledge management and organizational memory. Journal of Management Information Systems, 16(3), 103–117. \url{https://doi.org/10.1080/07421222.1999.11518258}

\hangindent=1.3em\hangafter=1
Zins, C. (2007). Conceptual approaches for defining data, information, and knowledge. Journal of the American Society for Information Science and Technology, 58(4), 479–493. \url{https://doi.org/10.1002/asi.20508}

\hangindent=1.3em\hangafter=1
Zwanzig, S., and Ahmad, N. (2024). Bayesian Inference: Theory, Methods, Computations. Chapman \& Hall/CRC.

\endgroup

\clearpage
\appendix
\section{Targeted Review of Bayesian Reference Books}\label{app:review}
This appendix gives the editions, passages and term counts behind the targeted review of Section~\ref{sec:bayes-books} and the four findings in Figure~\ref{fig:bayes-lit}.

\subsection{Method}\label{app:review-method}
Twelve books were examined in the editions listed in Table~\ref{tab:review-editions}. Their full texts were searched for the terms in Table~\ref{tab:review-counts}, and the passages found for the definitional questions were read. Page numbers are pages of the PDF files, which may differ from the print editions, except for Berger (1985), whose PDF is a scan of the print edition and whose pages are given as printed. Its counts come from text recognition of the scan and are less exact than the others. Counts come from text searches after joining line breaks, so they show patterns of use rather than exact figures. As stated in Section~\ref{sec:bayes-books}, frontier AI tools supported the searches, and the author reviewed the interpretations in context.

\begin{table}[H]
\centering
\caption{Editions examined in the targeted review.}
\label{tab:review-editions}
\small
\renewcommand{\arraystretch}{1.18}
\begin{tabularx}{\linewidth}{@{}>{\raggedright\arraybackslash}p{0.42\linewidth}>{\raggedright\arraybackslash}X@{}}
\toprule
\textbf{Book} & \textbf{Edition examined} \\
\midrule
Bayesian Data Analysis (Gelman et al., 2013) & Third edition, electronic edition, PDF of 22 January 2025, \url{http://www.stat.columbia.edu/~gelman/book/BDA3.pdf} \\
Bayesian Workflow (Gelman et al., 2026) & Authors' free electronic edition, PDF of 25 August 2026, \url{https://avehtari.github.io/Bayesian-Workflow/} \\
Statistical Decision Theory and Bayesian Analysis (Berger, 1985) & Second edition, Springer, scan of the print edition \\
Information Theory, Inference, and Learning Algorithms (MacKay, 2003) & Author's free edition, \url{https://www.inference.org.uk/itprnn/book.pdf} \\
Gaussian Processes for Machine Learning (Rasmussen and Williams, 2006) & Free edition, \url{https://gaussianprocess.org/gpml/chapters/RW.pdf} \\
Pattern Recognition and Machine Learning (Bishop, 2006) & Free edition from Microsoft Research \\
Probabilistic Machine Learning: An Introduction (Murphy, 2022) & Author's free online edition, version of April 2025 \\
Probabilistic Machine Learning: Advanced Topics (Murphy, 2023) & Author's free online edition, version of December 2025 \\
Bayesian Reasoning and Machine Learning (Barber, 2012) & Author's free online edition, version of March 2025 \\
Bayesian Inference: Theory, Methods, Computations (Zwanzig and Ahmad, 2024) & Chapman \& Hall/CRC edition \\
Bayesian Optimization (Garnett, 2023) & Author's free edition, PDF of 15 May 2024, \url{https://bayesoptbook.com/} \\
Algorithms for Decision Making (Kochenderfer et al., 2022) & Authors' free edition, PDF of 21 September 2026, \url{https://algorithmsbook.com/} \\
\bottomrule
\end{tabularx}
\end{table}

\begin{table}[H]
\centering
\caption{Occurrences of each term in the full text of each book. Counts depend on text extraction and show patterns of use. In Murphy (2023), two matches for ``wisdom'' are the surname of a cited author and are not counted. Counts for Berger (1985) come from text recognition of a scan.}
\label{tab:review-counts}
\small
\setlength{\tabcolsep}{5pt}
\renewcommand{\arraystretch}{1.18}
\begin{tabular}{@{}lrrrrrr@{}}
\toprule
\textbf{Book} & \textbf{\shortstack{prior\\distribution}} & \textbf{\shortstack{prior\\information}} & \textbf{\shortstack{prior\\knowledge}} & \textbf{\shortstack{inductive\\bias}} & \textbf{\shortstack{decision\\theory}} & \textbf{wisdom} \\
\midrule
Gelman et al. (2013) & 986 & 84 & 18 & 0 & 12 & 0 \\
Gelman et al. (2026) & 181 & 63 & 7 & 0 & 3 & 1 \\
Berger (1985) & 138 & 126 & 11 & 0 & 108 & 0 \\
MacKay (2003) & 33 & 1 & 1 & 0 & 20 & 0 \\
Rasmussen and Williams (2006) & 15 & 5 & 7 & 2 & 12 & 0 \\
Bishop (2006) & 79 & 0 & 3 & 0 & 24 & 0 \\
Murphy (2022) & 7 & 0 & 7 & 6 & 51 & 0 \\
Murphy (2023) & 24 & 2 & 11 & 12 & 22 & 0 \\
Barber (2012) & 10 & 3 & 3 & 0 & 8 & 0 \\
Zwanzig and Ahmad (2024) & 38 & 6 & 1 & 0 & 36 & 0 \\
Garnett (2023) & 21 & 0 & 6 & 0 & 59 & 1 \\
Kochenderfer et al.\ (2022) & 4 & 1 & 3 & 0 & 6 & 0 \\
\bottomrule
\end{tabular}
\end{table}

\subsection{Findings}\label{app:review-findings}
The findings follow the numbering of Figure~\ref{fig:bayes-lit}. Its implication is argued in Section~\ref{sec:bayes-books}.

\textbf{Finding 1: the prior distribution is defined, prior information is defined in one book and prior knowledge in none.} Most of the books keep the prior distribution apart from what is known. Gelman et al.\ (2013) state that ``the state of knowledge about anything unknown is described by a probability distribution'' (p.~21), recommend ``putting real prior information into the prior distribution'' (p.~197) and describe a weakly informative prior as providing information ``intentionally weaker than whatever actual prior knowledge is available'' (p.~65). Berger (1985) is the only book that defines prior information: ``This is information about $\theta$ arising from sources other than the statistical investigation. Generally, prior information comes from past experience about similar situations involving similar $\theta$'' (p.~2). He treats the prior distribution as one way of expressing it: ``A useful way of talking about prior information is in terms of a probability distribution on $\Theta$'' (p.~4). The definition covers information about the parameter $\theta$ only. Berger also uses information about observations, when the investigator can ``quantify subjective information about the marginal distribution of $X$'' (p.~95), but only as a means to construct the prior distribution on $\theta$. Information that shapes the data model $f(x\mid\theta)$, which Gelman et al.\ (2026, p.~497) count as prior information, falls outside the definition. None of the twelve books defines ``prior knowledge''. Berger, too, uses it as an everyday expression, as when the investigator ``may have considerable prior knowledge about the situation'' (p.~7). In the other eleven books, the expressions that appear are everyday expressions, and in the books that use both, the passages read use them interchangeably.

\textbf{Finding 2: prior context enters through the model as well as through the prior distribution.} Murphy (2022) states that ``a set of assumptions (also called inductive bias) that works well in one domain may work poorly in another'' and that ``The best way to pick a suitable model is based on domain knowledge'' (p.~43). Rasmussen and Williams (2006) note that the probabilistic approach can ``incorporate prior knowledge on the function and/or noise model'' (p.~55). Bishop (2006) writes that ``we may have prior knowledge that can be conveniently expressed through the prior distribution'' (p.~137), and elsewhere that ``This prior knowledge can be captured using the Markov random field model'', a model structure (pp.~407--408). Gelman et al.\ (2026) state that ``Prior information can enter a statistical analysis in many places'' and that ``the specification of the data model represents prior information or assumptions, whether this be linearity, a particular functional form such as the logistic, or the structure of a neural network or Gaussian process'' (p.~497). The same page adds study design: ``Design and sample size of a study will be chosen based on some balance of cost and prior assessment of possible effect sizes.'' Barber (2012) writes that ``Prior information about the structure of the data is often most naturally specified through a generative model'' (p.~336), and Murphy (2023) calls a constraint on the order of sequence labels ``prior information'' encoded in a model, with ``similar grammatical rules'' (p.~222). Kochenderfer et al.\ (2022) encode prior context in the initial state of an algorithm: ``prior knowledge of the transition dynamics can be used to initialize the counts differently'' (p.~348), ``Prior knowledge can be incorporated into Monte Carlo tree search in how we initialize $N$ and $Q$'' (p.~481) and ``If we have some prior information about where the agent might be in the state space, we can encode this in the initial belief'' (p.~401).

\textbf{Finding 3: within one book, one expression can carry different meanings, and different expressions the same meaning.} On page 497, Gelman et al.\ (2026) use ``prior information'' and, in the next paragraph, ``Prior knowledge also enters into non-Bayesian methods'' for the same idea. Zwanzig and Ahmad (2024) keep the two apart when they describe how to ``implement subjective information into a prior distribution'' (p.~45), but elsewhere use ``prior information'' for the prior distribution itself: the posterior kernel ``includes the prior information ($\pi$) and the information from the data'' (p.~26), where $\pi$ denotes the prior distribution. Berger's examples of prior information (1985, p.~2) are about different things under the one expression: results from other studies, ``a great deal of information available about $\theta_1$ and $\theta_2$ from different but similar pain relievers'', and an established belief, ``our prior opinion that this prediction is impossible (barring a belief in extrasensory perception)''.

\textbf{Finding 4: decisions are treated as a step distinct from inference, while Wisdom is not treated as a category.} Berger (1985) names two kinds of information beyond the data. ``The first is a knowledge of the possible consequences of the decisions'', which ``can be quantified by determining the loss'' (p.~1), and the second is prior information (p.~2). He adds that ``Bayesian analysis and decision theory go rather naturally together, partly because of their common goal of utilizing nonexperimental sources of information'' (p.~3). Knowledge of consequences, the input closest to judgments about value, is classed there as a kind of information, not as a separate category. Gelman et al.\ (2013) state that ``Decision analysis is inherently more complicated than statistical inference because it involves optimization over decisions as well as averaging over uncertainties'' (p.~248). MacKay (2003) adds actions and a utility function to the probability model (chapter 36). Murphy (2022) gives decision theory its own chapter (chapter 5) and Rasmussen and Williams (2006) its own section (section 2.4), and Bishop (2006) divides problems into ``two separate stages, the inference stage'' and the decision stage (section 1.5.4). Gelman et al.\ (2026) list the stages of a statistical workflow and end ``and then there is decision analysis'' (p.~188). Garnett (2023) states what the decision step adds: ``Bayesian inference enables us to express this uncertainty, but effective decision making additionally requires us to establish preferences over outcomes and act to maximize those preferences'' (p.~28). The accounts of difficulty differ: Bishop (2006) writes that ``the decision stage is generally very simple, even trivial, once we have solved the inference problem'' (p.~58), and MacKay (2003) calls decision theory ``trivial, apart from computational details'' (p.~463), while Gelman et al.\ (2013) call it more complicated than inference. Both can hold: once the costs and benefits of each outcome are fixed, computing the best action is straightforward, while choosing them and judging whether the model fits the problem is not. Zwanzig and Ahmad (2024) use decision theory differently, ``to give a deeper understanding of Bayesian inference'' (p.~89), judging inference methods themselves. Gelman et al.\ (2026) also describe decision theory as ``a tool for identifying incoherence'', whose purpose ``is to be able to go back and find problems with our underlying assumptions'' (p.~21). Kochenderfer et al.\ (2022), whose subject is decision making under uncertainty, model ``the preferences of an agent as a real-valued function over uncertain outcomes'' and introduce ``the maximum expected utility principle as a definition of rationality'' (p.~133). They state that ``Decision theory is a normative theory, which is prescriptive, not a descriptive theory, which is predictive of human behavior'' and that ``Human judgment and preference often do not follow the rules of rationality'' (p.~144). Judgment appears there as human behavior that departs from the normative rules, not as a category of its own.

The word ``wisdom'' appears only in its everyday sense, once in Gelman et al.\ (2026), ``There is wisdom in this anecdote'' (p.~478), and once in Garnett (2023, p.~128). The DIKW hierarchy does not appear in any of the twelve books.

\textbf{Evidence as a technical term.} ``Evidence'' denotes the marginal likelihood in several of the books. MacKay (2003) writes Bayes' theorem with the evidence as denominator (p.~41), Murphy (2022) defines ``the marginal likelihood or evidence for a model'' (p.~165), and Bishop (2006) notes that ``the model evidence is sometimes also called the marginal likelihood'' (p.~182). This is the technical sense distinguished from the epistemic sense in Section~\ref{sec:bayes-books}.

\subsection{Scope}\label{app:review-scope}
Jaynes, Probability Theory: The Logic of Science, could not be examined, because no complete official copy was available for download. The standard reference on Bayesian decision theory by Robert was not available, and neither were two textbooks on decision analysis, Smith's Bayesian Decision Analysis and Parmigiani and Inoue's Decision Theory. The finding that the expressions are used interchangeably rests on the passages read, not on every use of each term. The review is a targeted textual analysis, as stated in Section~\ref{sec:bayes-books}, not an exhaustive survey of Bayesian scholarship.

\end{document}